\documentclass[10pt,twocolumn]{IEEEtran}
\usepackage[T1]{fontenc}
\usepackage[utf8]{inputenc}
\usepackage{mathrsfs}
\usepackage{pifont}
\usepackage{bbding}
\usepackage{tipa}
\usepackage{color}
\usepackage{cite}
\usepackage{graphicx}
\usepackage{url}
\usepackage{amsfonts}
\usepackage{multicol}
\usepackage{float}
\usepackage{times}
\usepackage{psfrag}
\usepackage{subfigure}
\usepackage{stfloats}
\usepackage{footnote}
\usepackage{array}
\usepackage{amsmath,epsfig}
\usepackage{stmaryrd}
\usepackage{amssymb}
\usepackage{epic}
\usepackage{curves}
\usepackage{balance}
\usepackage{algorithm}
\usepackage{algpseudocode}
\usepackage{multirow}
\usepackage{ntheorem}
\usepackage{xcolor}
\usepackage{epstopdf}
\usepackage{enumerate}
\usepackage{lettrine}
\usepackage{threeparttable}
\usepackage{bm}
\usepackage{textcomp}

\def\BibTeX{{\rm B\kern-.05em{\sc i\kern-.025em b}\kern-.08em
    T\kern-.1667em\lower.7ex\hbox{E}\kern-.125emX}}
\begin{document}
\bstctlcite{IEEEexample:BSTcontrol}
\newtheorem{theorem}{Theorem}
\newtheorem{assumption}{Assumption}
\newtheorem{proposition}{Proposition}
\newtheorem{definition}{Definition}
\newtheorem{lemma}{Lemma}
\newtheorem{corollary}{Corollary}
\newtheorem{remark}{Remark}
\newtheorem{construction}{Construction}
\newtheorem{problem}{Problem}
\newtheorem{alg}{Algorithm}[section]

\newcommand{\supp}{\mathop{\rm supp}}
\newcommand{\sinc}{\mathop{\rm sinc}}
\newcommand{\spann}{\mathop{\rm span}}
\newcommand{\essinf}{\mathop{\rm ess\,inf}}
\newcommand{\esssup}{\mathop{\rm ess\,sup}}
\newcommand{\Lip}{\rm Lip}
\newcommand{\sign}{\mathop{\rm sign}}
\newcommand{\osc}{\mathop{\rm osc}}
\newcommand{\R}{{\mathbb{R}}}
\newcommand{\Z}{{\mathbb{Z}}}
\newcommand{\C}{{\mathbb{C}}}
\newcommand*{\affaddr}[1]{#1} 
\newcommand*{\affmark}[1][*]{\textsuperscript{#1}}
\newcommand*{\email}[1]{\texttt{#1}}

\title{Federated Learning with Unreliable Clients: Performance Analysis and Mechanism Design}
\author{Chuan Ma,~\IEEEmembership{Member,~IEEE,}
        Jun~Li,~\IEEEmembership{Senior Member,~IEEE,}\\
        Ming~Ding,~\IEEEmembership{Senior Member,~IEEE,}
        Kang~Wei,~\IEEEmembership{Student Member,~IEEE,}\\
        Wen~Chen,~\IEEEmembership{Senior Member,~IEEE,}
        and~H.~Vincent~Poor,~\IEEEmembership{Life~Fellow,~IEEE}
\IEEEcompsocitemizethanks{
\IEEEcompsocthanksitem This work was supported in part by the National Natural Science Foundation of China under Grant No. 61872184, 62002170, 62071296, in part by National key project 2020YFB1807703, 2018YFB1801102, STCSM 20JC1416502, and in part by U.S. National Science Foundation Grant CCF-1908308. (\emph{Corresponding author: Jun Li.})
\IEEEcompsocthanksitem Chuan Ma, Jun~Li and Kang~Wei are with School of Electrical and Optical Engineering, Nanjing University of Science and Technology, Nanjing, China. E-mail: \{kang.wei, jun.li, chuan.ma\}@njust.edu.cn.
\IEEEcompsocthanksitem Ming~Ding is with Data61, CSIRO, Sydney, Australia. E-mail: ming.ding@data61.csiro.au.
\IEEEcompsocthanksitem Wen~Chen is with the Department of Electronics Engineering, Shanghai Jiao Tong
University, Shanghai 200240, China. E-mail: wenchen@sjtu.edu.cn.
\IEEEcompsocthanksitem H.~Vincent~Poor is with Department of Electrical Engineering, Princeton University, NJ, USA. E-mail: poor@princeton.edu.}}

\maketitle

\begin{abstract}
Owing to the low communication costs and privacy-promoting capabilities, Federated Learning (FL) has become a promising tool for training effective machine learning models among distributed clients. However, with the distributed architecture, low quality models could be uploaded to the aggregator server by unreliable clients, leading to a degradation or even a collapse of training.
In this paper, we model these unreliable behaviors of clients and propose a defensive mechanism to mitigate such a security risk. Specifically, we first investigate the impact on the models caused by unreliable clients by deriving a convergence upper bound on the loss function based on the gradient descent updates. Our theoretical bounds reveal that with a fixed amount of total computational resources, there exists an optimal number of local training iterations in terms of convergence performance. We further design a novel defensive mechanism, named deep neural network based secure aggregation (DeepSA).
Our experimental results validate our theoretical analysis. In addition, the effectiveness of DeepSA is verified by comparing with other state-of-the-art defensive mechanisms.
\end{abstract}

\section{Introduction}
Machine learning (ML) technologies, e.g., deep learning, have revolutionized the ways that information is extracted with ground breaking successes in various areas. Meanwhile, owing to the advent of the Internet of things (IoT), the number of intelligent applications with edge computing, such as smart manufacturing, intelligent transportation and intelligent logistics, is growing exponentially~\cite{chiang2016fog,Li2019Contract,S2020Privacy,Shaham2020Privacy,9200330}. As such, the conventional centralized deep learning is no longer capable of efficiently processing the dramatically increased data from the vast IoT devices. To tackle this challenge, distributed learning frameworks have emerged, e.g., federated learning (FL), enabling the decouple of data provision by distributed clients and aggregating ML models at a centralized server~\cite{vepakomma2018no,McMahan2017Communication,Li2020Fed}. Through local training and central aggregating iteratively, FL does not require clients to share their sensitive data with the central server, thereby effectively reducing transmission overheads as well as preserving clients' privacy to some extent~\cite{yu2017survey,ma2020safeguarding,Wei2020Fed}.

Although the clients' data are not explicitly exposed in the original format, it is still possible for adversaries to infer clients' private information approximately, especially when the architecture of the FL model and its parameters are not completely protected.
Moreover, the existence of unreliable clients may further incur security issues in IoT applications. This is because the server in a FL system has no access to the clients' data, nor does it have a full control of the clients' behaviors. As a consequence, a client may deviate from the normal behaviors during the course of FL, which is termed as \textbf{unreliable client} in this work. {\color{black}Unreliable behaviors may be caused intentionally, e.g., by a malicious attacker disguised as a normal client, or unintentionally, e.g., by a client with hardware and/or software limitaions/defects in IoT.}  For example, in smart manufacturing scenarios, engines with sensors that have abnormal traffic and irregular reporting frequency may cause industrial production interruption thus resulting in huge economic losses for factories \cite{8769942,7723838}.

Unreliable clients in FL, for example, may manipulate their outputs sent to the server and they can dominate the training process and change the judging boundary of the global model, or make the global model deviate from the optimal solution. To model these clients,
the work in~\cite{baruch2019little} proposed that an unreliable client may interfere with the process of FL by applying limited changes to the uploaded model parameters.
The work in~\cite{Bagdasaryan2020How} proposed a model-replacement method that demonstrated its efficacy on poisoning models of standard FL tasks in IoT. In addition, this work also developed and evaluated a generic constrain-and-scale technique that incorporates the evasion of defensive mechanism into the abnormal clients' loss function during training.
{\color{black}Therefore, how to design defensive algorithms against abnormal clients in FL becomes crucial.}
In order to detect abnormal updates in FL, the work of~\cite{9066920} applied the results of client-side cross-validation for reducing the weights of bad updates when performing aggregation, where each update is evaluated over other clients' local data. Similarly, the work in~\cite{DBLP:journals/corr/abs-1912-11464} also focused on the weights and they presented a novel aggregation algorithm with the residual-based re-weighting method.
The work in~\cite{Zhao2020Privacy} considered the existence of unreliable participants and utilized an auxiliary validation data to compute a utility score for each participant to reduce the impact of these unreliable participants, while the work in~\cite{Zhao2020PDGAN} directly removed the corresponding model parameters from the training procedure if the accuracy of client is lower than a predefined threshold.
The work in~\cite{Mu2019Byzantine} proposed a robust aggregation rule, called adaptive federated averaging, which detects and discards malicious or bad local updates based on a hidden Markov model.
The work in~\cite{fang2019local} performed the first systematic study on local model poisoning attacks on FL, in which they formulate attacks as optimization problems and test four different robust FL methods.
However, all of these works lack theoretical analysis on the performance of FL systems with the existence of unreliable clients.

It should be noted that the analysis and optimization for the basic FL system has already been investigated in~\cite{zhao2018federated, zhou2018On, Wang2019Adaptive, stich2018local}, yet there are no analytical results on the security aspects in a FL system. Therefore, in this work, we conduct analysis in the context of FL with unreliable clients. We first introduce the unreliable model of clients in FL systems and derive the theoretical convergence bounds. Through our theoretical results, we find that there exists an optimal local training iteration that leads to a best system performance within a constraint on total computing resources. Then, we design a novel defensive mechanism, referred to as deep neural network (DNN) based secure aggregation (DeepSA), to efficiently reduce the negative effects caused by unreliable clients.

The major contributions of this paper can be summarized as follows:
\begin{itemize}
\item We involve unreliable clients in FL, which model parameters will be scaled down and noised before uploading. Further, we derive the upper bound of loss function in FL systems with a given computational resources. Our theoretical bound reveals that there exists an optimal number of local training epochs to achieve the best convergence performance.
\item We propose a novel defensive mechanism, i.e., DNN-based DeepSA, which can detect abnormal models, and then alleviates the negative effects by removing them from the aggregation.
\item We conduct extensive experiments on the proposed model with the multi-layer perceptron model and real-life dataset. Our experimental results are shown to be consistent with the theoretical ones. Also, compared with other existing defensive algorithms, the proposed one can improve the FL model performance effectively.
\end{itemize}

The rest of this paper is organized as follows. Section~\ref{sec:prelim} introduces background of FL. Section~\ref{sec:threat_model} details the system models. In Section~\ref{sec:conver} we analyzes convergence bounds of FL system with unreliable clients. Section~\ref{sec:DeepSA} proposes the DeepSA algorithm to address the unreliable problem, and the experimental results are shown in Section~\ref{sec:experi}. Finally, we conclude the paper in Section~\ref{sec:concl}. In addition, a summary of notation is listed in Table~\ref{tab}.
\begin{table*}
\centering
\caption{Summary of main notation}\label{tab}
{
\begin{tabular} {|c|c|}
  \hline
  Notation& Description\\
  \hline
  $\boldsymbol{w}_{i}^{(k\tau)}$ & The uploaded model of $i$-th client at the $k$-th communication round\\
  \hline
  $\tau$ & The number of local training epochs\\
  \hline
  $F_{i}(\cdot)$ & The local objective function of the $i$-th client\\
  \hline
  $\widehat{\boldsymbol{w}}_{i}^{(k\tau)}$ & The local model of the $i$-th unreliable client \\
  \hline
  $\alpha$ & The scaling factor with range [-1,1]\\
  \hline
  $\boldsymbol{n}$ & The additive noise\\
  \hline
  $p_{\textrm{U}}$ & The probability of the unreliable behavior\\
  \hline
  $p_{i}$ & The aggregating weight based on the training data size of $\mathcal D_i$\\
  \hline
  $T$ & The number of total training iterations\\
  \hline
  $k$ & The number of total aggregation \\
  \hline
\end{tabular}}
\end{table*}

\section{Preliminaries For Federated Learning}\label{sec:prelim}
In this section, we will introduce the basic concepts of FL. As a kind of distributed training frameworks~\cite{McMahan2017Communication}, FL can promote user privacy by its unique distributed learning mechanism. In FL, all clients share the same learning objective and model structure, where a central server sends the current global model parameters $\boldsymbol w$ to all clients $\mathcal{C}_i$, $\forall i \in \mathcal{M}\triangleq \left\{1,2,\ldots,M\right\}$, in each communication round. Then, all clients update local models based on the shared global model and local data set $\mathcal{D}_{i}$. After local training, all local models will be uploaded to the server by clients, and then aggregated by the server as the current global model, which is expressed as
\begin{equation}\label{equ:aggregation}
\boldsymbol{w}^{(k\tau)}=\sum\limits_{i\in\mathcal{M}}p_{i}\boldsymbol{w}_{i}^{(k\tau)},
\end{equation}
where $\boldsymbol{w}_{i}^{(k\tau)}$ is the uploaded model of $i$-th client at the $k$-th communication round, $\boldsymbol{w}^{(k\tau)}$ is the global model after aggregation at the $k$-th communication round, $\tau$ is the number of local training epochs, and $p_{i}=\vert \mathcal{D}_{i}\vert/\vert \mathcal{D}\vert$ is the aggregating weight based on the size of $\mathcal D_i$, where $\mathcal{D}=\sum_{i\in\mathcal{M}}\cup \mathcal{D}_{i}$ and $|\cdot|$ represents the cardinality of a set, respectively. At the server side, the goal is to learn a model over data that resides at the $M$ associated clients. Formally, this FL task can be expressed as
\begin{equation}\label{equ:global_loss}
\boldsymbol{w}^{*}=\mathop{\arg\min}_{\boldsymbol{w}}\sum_{i\in\mathcal{M}}{p_{i}F_{i}(\boldsymbol{w})},
\end{equation}
where $F(\boldsymbol{w})=\sum_{i\in \mathcal{M}}p_{i}F_{i}(\boldsymbol{w})$ and $F_{i}(\cdot)$ is the local objective function of the $i$-th client.

In addition, the FL in this paper adopts the optimization method of gradient descent. In order to capture the divergence between the gradient of a local and global loss function, the gradient divergence is defined as follows~\cite{Wang2019Adaptive}.
\begin{definition}
For any $i\in \mathcal{M}$ and $\boldsymbol{w}$, an upper bound of $\Vert\nabla F_{i}(\boldsymbol{w})-\nabla F(\boldsymbol{w})\Vert$ is defined as $\delta_{i}$, i.e.,
\begin{equation}
\Vert\nabla F_{i}(\boldsymbol{w})-\nabla F(\boldsymbol{w})\Vert\leq \delta_{i}.
\end{equation}
We also define $\delta\triangleq\frac{\sum_{i\in \mathcal{M}}{\vert\mathcal{D}_{i}\vert\delta_{i}}}{\vert\mathcal{D}\vert}$. If the size of each local dataset is same, we know that $\delta=\frac{1}{M}\sum_{i=1}^{M}\delta_{i}$.
\end{definition}
This divergence is governed by how the data is distributed at different clients.

\begin{figure}
\centering
\includegraphics[width=3in,angle=0]{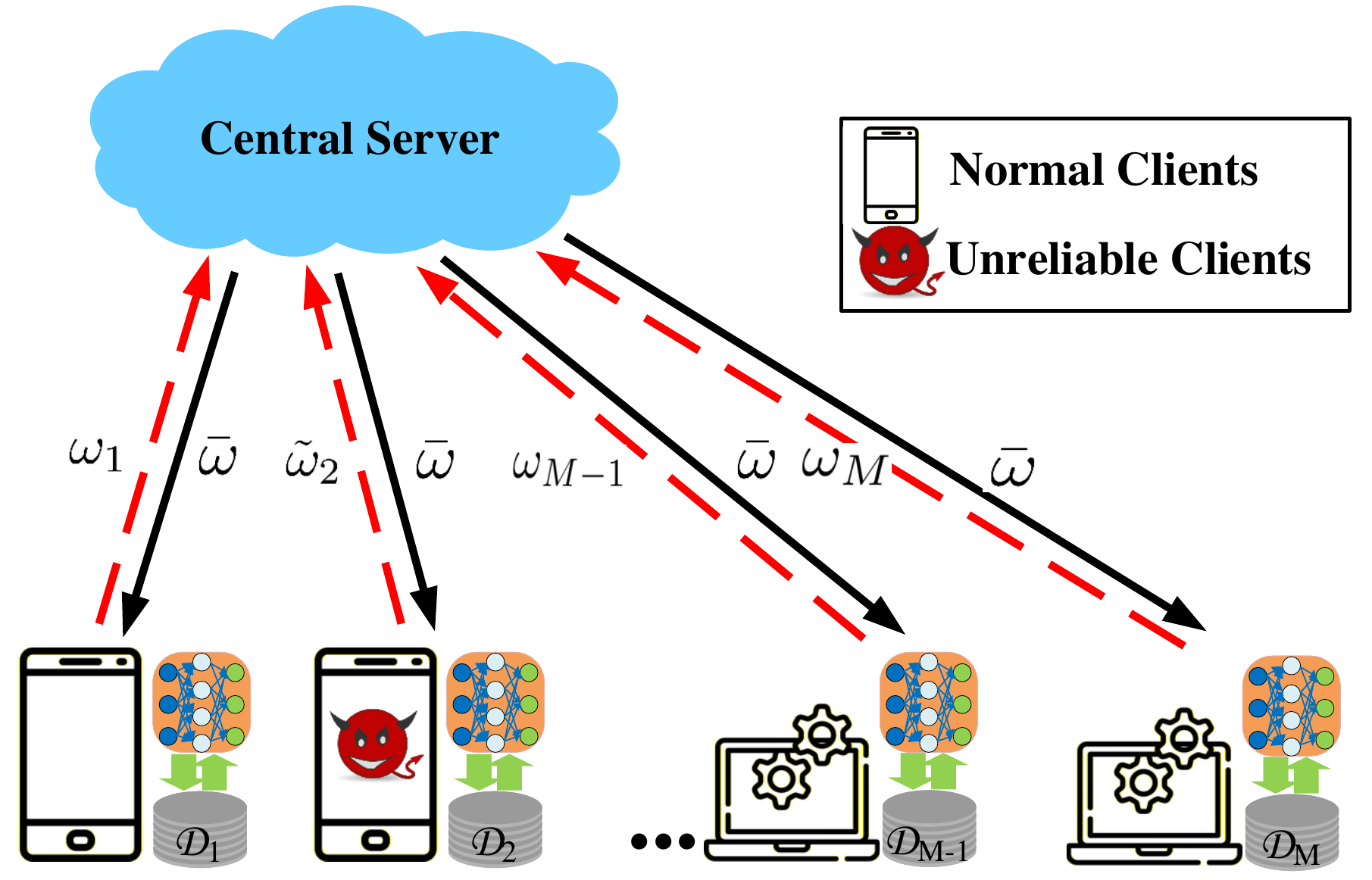}
\caption{A FL training framework with unreliable clients.}
\label{fig:FL Model}
\end{figure}
\section{System Models}\label{sec:threat_model}
In FL, the unreliable model updates might exist in a wireless transmission environment. Thus, these flawed uploads will impair the effectiveness of the global model, misleading the updated AI model away from optimality. To be more specific, abnormal behaviors generally can be classified into two categories: intentional and unintentional. Intentional adversary clients, also regarded as malicious clients, usually aim to sabotage the system performance or even destroy the learned model. For example, the values of the uploaded parameters may be scaled down or even completely reversed. By contrast, unintentional behaviors could happen without any particular purpose, for example, a noisy version of parameters could be uploaded to the server. In this paper we propose to model both types of abnormal clients.

We consider a FL system consisting of a single central server and $M$ clients, as shown in Fig. \ref{fig:FL Model}. We assume each clients may upload unreliable models throughout the whole training process.

\subsection{Adversary Model}
Each client is assumed to have a local model with the same structure, and corresponding model parameters uploaded for each epoch are of the same format. The shared model is guided by these parameter vectors to the optimal value. Then, we denote by $\widehat{\boldsymbol{w}}_{i}^{(k\tau)}$ the local model of the $i$-th unreliable client, and express as
\begin{equation} \label{abnormal}
\widehat{\boldsymbol{w}}_{i}^{(k\tau)} = \alpha\boldsymbol{w}_i^{(k\tau)} + \boldsymbol{n}_i^{(k\tau)},
\end{equation}
where $\alpha \in [-1,1]$ denotes the scaling factor and $\boldsymbol{n}$ denotes the additive noise and is assumed to follow a Gaussian distribution with $N(0,\sigma^2)$.
Eq.~(\ref{abnormal}) can well capture the abnormal behaviors as the scalar is used to model the malicious clients and the random noise denotes the undesirable perturbation on the uploaded models. For example, when $\alpha=-1$, it means an adversarial client will completely reverse the uploaded parameters on purpose, and can be recognized as a malicious behavior.
Due to the aggregation process in Eq.~\eqref{equ:aggregation}, we have
\begin{equation}\label{equ:reversed_model}
\begin{split}
\bar{\boldsymbol{w}}^{(k\tau)}&=\sum\limits_{i \in M} p_i \tilde{\boldsymbol{w}}_i,
\end{split}
\end{equation}
where 
\begin{equation}\label{ww}
\begin{split}
\tilde{\boldsymbol{w}}_i&=
            \begin{cases}
              \boldsymbol{w}_i, & \text{with probability $P_{\textrm{U}}$;} \\
              \alpha\boldsymbol{w}_i + \boldsymbol{n}_i, & \text{with probability $1-P_{\textrm{U}}$,}
            \end{cases}\\
            \end{split}
\end{equation}
{\color{black}and $p_{\textrm{U}}$ denotes the probability of the unreliable behavior}\footnote{{\color{black}We assume a same value of $p_{\textrm{U}}$ for all clients in this work. Different unreliable probabilities for different clients maybe out of the scope, and can be our future work.}}.

\section{Convergence Analysis}\label{sec:conver}
In this section, we will propose a theoretical analysis on the convergence of the FL system considering the existing of abnormal behaviors. For the purpose of facilitating the analysis, we make the following assumptions on the loss function.
\begin{assumption}\label{assum:converg_analy}
We assume the following conditions are satisfied for all $i$, $\forall i \in \mathcal{M}$:
\begin{enumerate}
\item[\emph{1)}] $F_{i}(\boldsymbol{w})$ is convex;
\item[\emph{2)}] All model parameters satisfy $\Vert\boldsymbol{w}\Vert\leq \Theta$;
\item[\emph{3)}] $F_{i}(\boldsymbol{w})$ is $\rho$-Lipschitz, i.e., $\Vert  F_{i}(\boldsymbol{w})- F_{i}(\boldsymbol{w}')\Vert\leq \rho\Vert \boldsymbol{w}-\boldsymbol{w}'\Vert$, for any $\boldsymbol{w}$, $\boldsymbol{w}'$;
\item[\emph{4)}] $F_{i}(\boldsymbol{w})$ is $\beta$-smooth, i.e., $\Vert \nabla F_{i}(\boldsymbol{w})-\nabla F_{i}(\boldsymbol{w}')\Vert\leq \beta\Vert \boldsymbol{w}-\boldsymbol{w}'\Vert$, for any $\boldsymbol{w}$, $\boldsymbol{w}'$;
\item[\emph{5)}] $\eta\leq\frac{1}{\beta}$, where $\eta$ is the step size;
\item[\emph{6)}] $\Vert F\left(\boldsymbol{w}^{t}\right)-F(\boldsymbol{w}^{*})\Vert\geq\varepsilon$, for all $\boldsymbol{w}$ during FL training.
\end{enumerate}
\end{assumption}
We also assume that the clients participating in the training hold the same amount of data, i.e. $p_{i}=\frac{1}{M}$. In general, these assumptions with some restrictions for the convenience of theoretical derivation can be satisfied.
\subsection{Convergence Analysis}
In this subsection, we evaluate the performance of FL under the abnormal behaviors by an upper bound on the difference between $\mathbb{E}\left\{F\left(\bar{\boldsymbol{w}}^{(T)}\right)\right\}$ and $F(\boldsymbol{w}^{*})$, where $\bar{\boldsymbol{w}}^{(T)}$ is the final global parameters of FL system containing $M$ potential unreliable clients and $\boldsymbol{w}^{*}$ is the optimal model parameters that minimizes $F(\boldsymbol{w})$.
\begin{theorem}\label{theorem:bound_reverse}
For some $\varepsilon>0$ and $\Theta>0$, when the clients in the FL system behave unreliable with probability $p_\emph{U}$, the convergence upper bound with a fixed total number of iterations $T$ is given by
\begin{equation}
\begin{split}
&\mathbb{E}\left\{F\left(\bar{\boldsymbol{w}}^{(T)}\right)\right\}-F(\boldsymbol{w}^{*})\\
&\leq\frac{1}{T\left(\omega\eta(1-\frac{\beta\eta}{2})-\frac{\rho \left(\phi(\tau)+\frac{{ {p_\emph{U}}}}{M}\left[ {(1 - \alpha )M\Theta  + \frac{{2\sqrt M \sigma }}{\pi }} \right]\right)}{\tau\varepsilon^{2}}\right)},\\
\end{split}
\end{equation}
where $\phi(\tau)=\frac{\delta}{\beta}\left((\eta\beta+1)^{\tau}-1\right)-\eta\delta \tau$, $\varphi=\omega(1-\frac{\beta\eta}{2})$, and $\omega \triangleq \min\limits_{k} \frac{1}{\Vert\boldsymbol{w}^{(k\tau)}-\boldsymbol{w}^{*}\Vert^{2}}$.\\
\end{theorem}
\begin{IEEEproof}
See Appendix~\ref{appendix:bound_reverse}.
\end{IEEEproof}

The upper bound given by~\textbf{Theorem~\ref{theorem:bound_reverse}} demonstrates the convergence result of the FL system with unreliable clients. A lower bound means that the value of the system loss function converges closer to the optimal one.

\subsection{Discussions on the Convergence Bound}
In this subsection, we will provide several key observations on the convergence bound.

\begin{proposition}\label{rem:reverse_1}
If there is no unreliable client, the convergence bound of FL increases, which also means a \textbf{worse} system performance, as the local epochs $\tau$ increases. Since other parameters are basically fixed, the influence of $\tau$ on this theoretical value is the most noteworthy.
\end{proposition}
\begin{IEEEproof}
When $p_{\textrm{U}}=0$, the convergence upper bound can be expressed as $1/{T\left(\omega\eta(1-\frac{\beta\eta}{2})-\frac{\rho \left(\phi(\tau)\right)}{\tau\varepsilon^{2}}\right)}$. It is evident that $\frac{\rho \left(\phi(\tau)\right)}{\tau\varepsilon^{2}}$ increases with an larger $\tau$ and leads to a larger bound.
\end{IEEEproof}
\begin{proposition}\label{cor:reverse_2}
When the probability of unreliable behaviors $p_{\emph{U}}$ is larger, the convergence performance becomes worse. However, when this percentage is fixed, the performance of the system will \textbf{improve} with the number of total clients $M$.
\end{proposition}

\begin{IEEEproof}
Considering the analytical part in the convergence bound related to $M$, we find the value $\frac{{ {p_\textrm{U}}}}{M}\left[ {(1 - \alpha )M\Theta  + \frac{{2\sqrt M \sigma }}{\pi }} \right]$ decreases with the increase of $M$ when $p_{\textrm{U}}$ is fixed, thus the convergence bound becomes smaller.
\end{IEEEproof}

We note that when the total number of iterations $T$ is constant, the local training iterations $\tau$ should be as small as possible if there is no unreliable client.
This is because that when $\tau = 1$, the FL system based on distributed gradient descent is equivalent to a centralized training model~\cite{Wang2019Adaptive}.
However, in an unreliable circumstance, there exists an optimal value of $\tau \in [1,T]$ ($T$ is an integer multiple of $\tau$) to have the optimal convergence performance. Therefore, we can make the following proposition.
\begin{proposition}\label{proposition:optim_tau}
Under the unreliable behaviors of clients with a fixed $T$, the convergence upper-bound is a convex function of the number of local epochs $\tau$, if we treat $\tau$ as a continuous variable.
\end{proposition}
\begin{IEEEproof}
See Appendix~\ref{appendix:optim_tau}.
\end{IEEEproof}

From~\textbf{Proposition~3}, we can see that there exists an optimal $\tau$ which can minimize the value of the loss function to obtain a satisfied learning performance.
\section{Defensive Mechanism Design}\label{sec:DeepSA}
In this section, we will use a crafted DNN to detect the existence of unreliable clients. {\color{black}Current defensive mechanisms, such as Krum \cite{NIPS2017_f4b9ec30} and Secprobe \cite{8825829}, usually need an online testing dataset to adjust the aggregation weight.} The testing dataset is either from clients, which may pose privacy issue, or using a public one that may affect the accuracy. Thus, in this work we consider training an offline detector to recognise the abnormal clients.  A basic binary anomaly detection technique using DNN operates in two steps. First, the DNN is trained on the normal training data to learn all normal labels. Second, each test instance is provided as an input to the DNN.
If the DNN accepts the test input, it is labeled as normal and if the network rejects a test input, it is an anomaly. Therefore, we propose the DeepSA algorithm based on a crafted DNN in one-class setting. The main implemental process is operated in the server with a new functional module. To complete this module, the detector is pre-trained before FL with several normal parameter inputs, and these parameters can be obtained from clients or a public dataset. Once the pre-training process ends, this module can be used for anomaly detection.

In FL, the set of local models received by the server at the $k$-th communication round can be expressed as
\begin{equation}
\boldsymbol{o}^{(k)}\triangleq\left\{\boldsymbol{w}^{(k\tau)}_{1}, \boldsymbol{w}^{(k\tau)}_{2},\ldots, \boldsymbol{w}^{(k\tau)}_{i},\ldots,\boldsymbol{w}^{(k\tau)}_{U}\right\}.
\end{equation}
We use a DNN based anomaly detector, denoted by $\mathfrak{D}$, which can be viewed as a classifier to assign a label (normal or abnormal). Typically, the outputs produced by this detector, are one of the following two types:
1) \emph{Scores}: Scoring techniques assign an detecting score to each instance, which is utilized to analyze the possibility of unreliable clients; 2) \emph{Labels}: Techniques in this category assign a label (benign or malicious) to each test instance. In our DNN based detector, we define the detecting result of a test instance (or observation) by $\boldsymbol{z}^{(k)}=[z_{1}^{(k)}, z_{2}^{(k)},\ldots,z_{U}^{(k)}] \in \left\{0,1\right\}^{U}$. If $z_{i}=0$, it represents that $\boldsymbol{w}^{(k\tau)}_{i}$ is the unreliable model. Therefore, the detecting process can be given by
\begin{equation}
\boldsymbol{z}^{(k)} = \mathfrak{D}\left(\boldsymbol{o}^{(k)}, \boldsymbol{o}^{(k-1)},\ldots,\boldsymbol{o}^{(1)}\right),
\end{equation}
where $\mathfrak{D}$ is the detector and we assume that whole sets of local models (from $1$-st to $k$-th communication round) can be utilized as the input of this detector.

In detail, the server will receive the observations $\boldsymbol{o}^{(k)}$ at the $k$-th communication round, and try to identify these abnormal models in them.
For a normal client, there should be a certain level of correlation among its uploaded parameters in consecutive communication rounds. However, manipulation of parameters by anomaly clients may break this correlation, hence the previous observations can also assist detecting abnormal models.
Therefore, in order to enhance this difference, we use an input reshaping approach to ensure the shift-invariance property for our detector, which can be expressed as
\begin{equation}
\boldsymbol{O}^{d}=
\left[
  \begin{array}{ccccc}
    \boldsymbol{w}_{1}^{(k\tau)} & \boldsymbol{w}_{1}^{((k-1)\tau)} & \ldots&\boldsymbol{w}_{1}^{((k-d)\tau)}\\
    \boldsymbol{w}_{2}^{(k\tau)} & \boldsymbol{w}_{2}^{((k-1)\tau)} &\ldots&\boldsymbol{w}_{2}^{((k-d)\tau)}\\
    \vdots & \vdots &\ddots&\vdots\\
    \boldsymbol{w}_{U-1}^{(k\tau)} & \boldsymbol{w}_{U-1}^{((k-1)\tau)} &\ldots&\boldsymbol{w}_{U-1}^{((k-d)\tau)}\\
    \boldsymbol{w}_{U}^{(k\tau)} & \boldsymbol{w}_{U}^{((k-1)\tau)} &\ldots&\boldsymbol{w}_{U}^{((k-d)\tau)}\\
  \end{array}
\right],
\end{equation}
where $d$ is the depth of the observation. The input $\boldsymbol{O}^{d}$ is shaped as a multi-dimensional vector ($U\times1\times d\times s_{w}$), where $s_{w}$ is the size of the standard uploaded model. With this input design, we will introduce the construction of our DNN based detector.

\begin{figure}[ht]
\centering
\includegraphics[width=3in,angle=0]{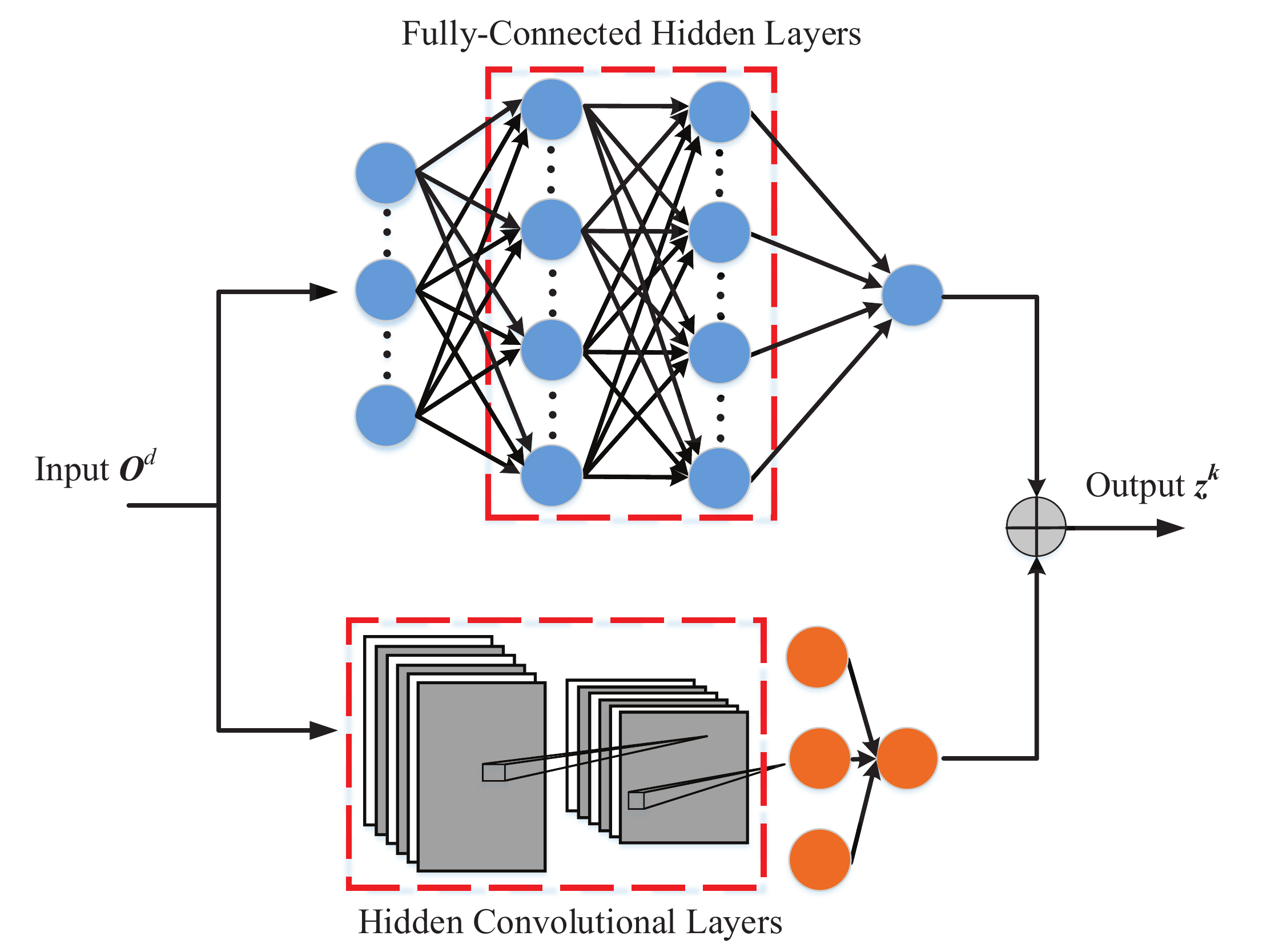}
\caption{The architecture of the DNN based detector.}
\label{fig:DNN_Architecture}
\end{figure}
As shown in Fig.~\ref{fig:DNN_Architecture}, the DNN detector consists of two parallel pipelines and achieves an output with the XOR operator. The input is $\boldsymbol{O}^{k}$ and the output is the symbol $\boldsymbol{z}^{k}$.
Using the fully-connected layers and sufficient training, we can identify the intentional behaviors, such as the scaling operation on the parameters.
In addition, for the randomized parameters, the convolutional layer can be more useful since the correlation between the noised and normal parameters is expected to be low.
In the convolutional layer, we use zero-padding with stride size $B$, and set the filter size to $B\times l$, where $l$ is the depth of the filter. This setting is based on the observation that each sub-vector is strongly correlated with $2B$ neighbouring sub-vectors due to the structure of uploaded models.
\begin{algorithm}
\caption{Secure Aggregation on the Server Side}
\label{alg:DeepSA}
\begin{algorithmic}[1]
\Require The observation of all uploaded models with $d$ depth $\boldsymbol{O}^{d}$, the well trained DNN based detector $\mathfrak{D}$
\Ensure The global model $\boldsymbol{w}^{(k\tau)}$
\State Wait for clients to upload their weights until there are $M$ clients' models $\boldsymbol{o}^{k}=\left\{\boldsymbol{w}^{(k\tau)}_{1}, \boldsymbol{w}^{(k\tau)}_{2},\ldots, \boldsymbol{w}^{(k\tau)}_{i},\ldots,\boldsymbol{w}^{(k\tau)}_{M}\right\}$
\State Update the input $\boldsymbol{O}^{d}$ with the fresh uploaded models $\boldsymbol{o}^{k}$
\State Obtain the detecting results with the DNN based detector $\mathfrak{D}$ as $\boldsymbol{z}^{k} = \mathfrak{D}\left(\boldsymbol{O}^{d}\right)$
\State Average the benign models and obtain $\boldsymbol{w}^{(k\tau)}=\sum_{i=1}^{M} z_{i}p_{i}\boldsymbol{w}^{(k\tau)}_{i}$
\State Send the new averaged model $\boldsymbol{w}^{(k\tau)}$ to all clients\\
\Return $\boldsymbol{w}^{(k\tau)}$
\end{algorithmic}
\end{algorithm}

After introducing this crafted DNN based detector, we present our proposed defence algorithm as shown in \textbf{Algorithm~\ref{alg:DeepSA}}. As discussed above, the existence of abnormal clients indicates that the parameters uploaded by them may be disruptive, and it may reduce the accuracy of the global model. To mitigate their effect on the model accuracy, we remove the malicious models in this communication rounds using our DNN based detector.

\textbf{Algorithm~\ref{alg:DeepSA}} gives the pseudocode of secure aggregation on the server side. The server first waits for the local model from each client. When all the clients finish uploading their models to the server, these models are utilized to update the input $\boldsymbol{O}^{d}$ with the fresh uploaded models. Then, the server can obtain the detecting results with the DNN based detector and average the benign models. For the whole system, reducing the number of clients is equivalent to reducing the amount of training data, which will reduce the generalization of the global model. However, compared with the damage brought by unreliable clients, these losses are acceptable. In addition, considering that even a reliable client may upload a model with poor quality, the decision result of DNN detector will only take effect in the current round of communication.
\section{Experimental Results}\label{sec:experi}
In this section, we first evaluate the performance of the analytical results with the unreliable clients, and verify the effectiveness with the experimental results. Then, we demonstrate the effectiveness of the proposed defensive mechanism by comparing with other algorithms\footnote{Related codes can be found in the following link: https://github.com/JJisbug/UnreliableClientsinFL.}.

\subsection{Experimental Settings}
\subsubsection{Dataset}
In our experimental results, we use four benchmark datasets for different tasks:
\begin{itemize}
  \item MNIST and Fashion-MNIST dataset, which both have $70K$ digit images of size $28 \times 28$, are split into $60K$ training and $10K$ test samples;
  \item CIFAR-10 dataset, which consists of $60K$ color images in 10 object classes such as deer, airplane, and dog with 6000 images included per class, are split into $50K$ training and $10K$ test samples;
  \item Adult dataset, which has around $32K$ tabular samples and each sample has 14 attributes, is split into $20K$ training and $12K$ test samples.
\end{itemize}

We consider the data distribution as independent identically distributed (i.i.d), i.e. clients in the FL system possess a same amount of data from training sets randomly and independently.
\subsubsection{Parameter Settings}
We use the multi-layer perceptron (MLP) as the training model to construct the FL system, and each client locally computes stochastic gradient descend (SGD) updates on each dataset, and then aggregate updates to train a globally shared classifier. We conduct three cases for the unreliable client to verify the analytical results as follows:
\begin{itemize}
  \item Case I: $\alpha=-1$ and $\sigma=0.1$, which an unreliable client uploads a completely inverse parameter with small noise;
  \item Case II: $\alpha=0.8$, and $\sigma=0.5$, which a large noise is added;
  \item Case III: $\alpha=0.5$, and $\sigma=0.3$, which the parameter is scaling half with a medium noise.
\end{itemize}
In addition, we set the total number of clients $M=50$ and the total learning iterations $T(k\tau)=300$. We run each experiment for 20 times and record the average results. If a client uploads unreliable parameters in all communication rounds, then this scenario can be treated as a special case in which we assume there are certain percentages of unreliable clients, and other clients will upload reliable parameters during the whole training process.
\subsubsection{Comparing defensive mechanisms}
To show the effectiveness of the proposed defensive mechanisms, we provide the following defensive mechanisms:
\begin{itemize}
  \item Krum \cite{NIPS2017_f4b9ec30}: the aggregated parameters are chosen according to the minimum geometric gradient rule.
  \item Secprobe \cite{8825829}: the aggregated parameters are chosen according to the testing accuracy.
  \item Pearson \cite{7783532}: the aggregated weights are adjusted by the Pearson correlation.
\end{itemize}
\subsection{Theoretical Results}
\begin{figure}
\centering
\subfigure[MNIST Dataset]{\label{con1}
  \includegraphics[width=0.22\textwidth]{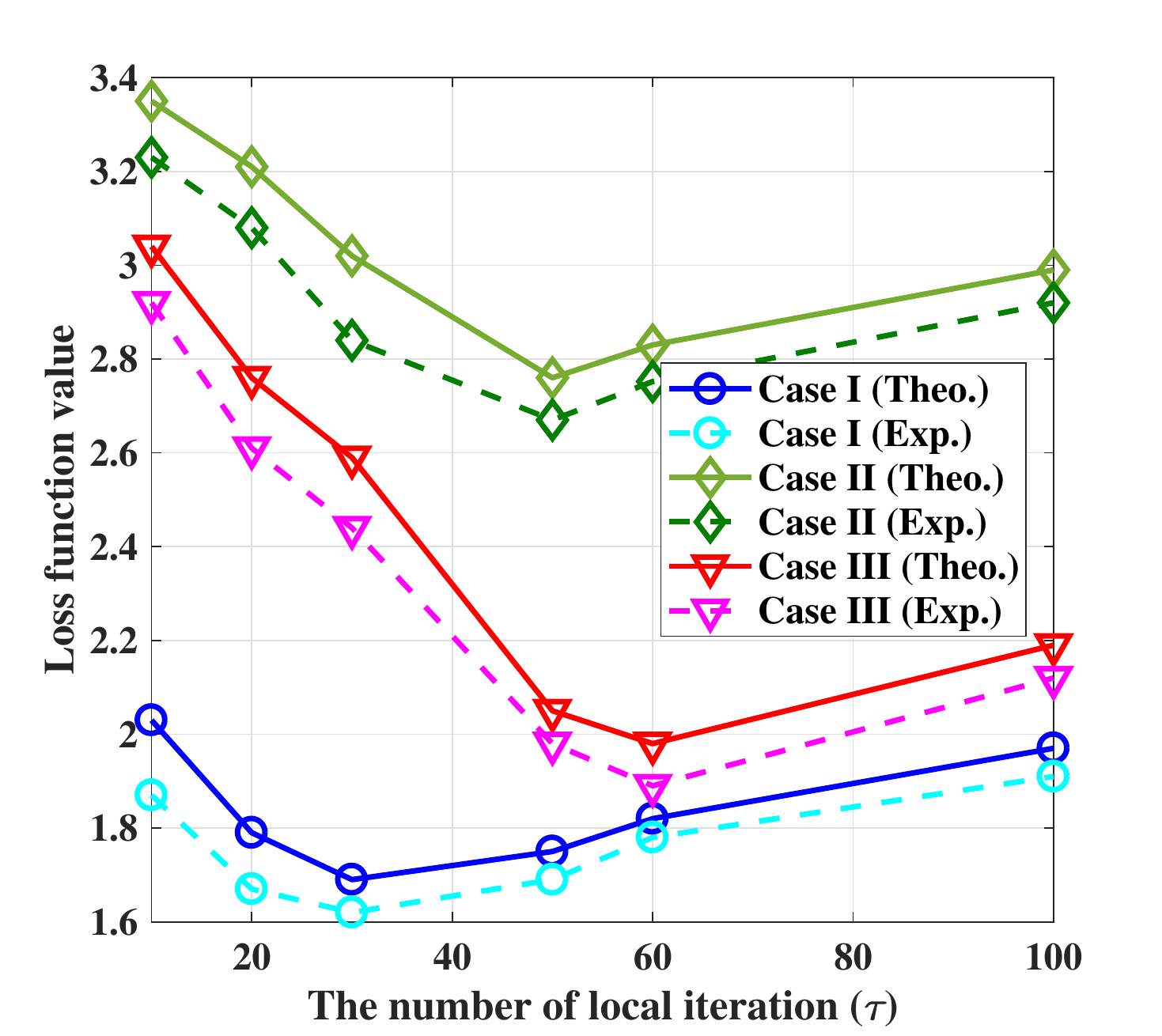}
}
\subfigure[Fashion-MNIST Dataset]{\label{con2}
  \includegraphics[width=0.22\textwidth]{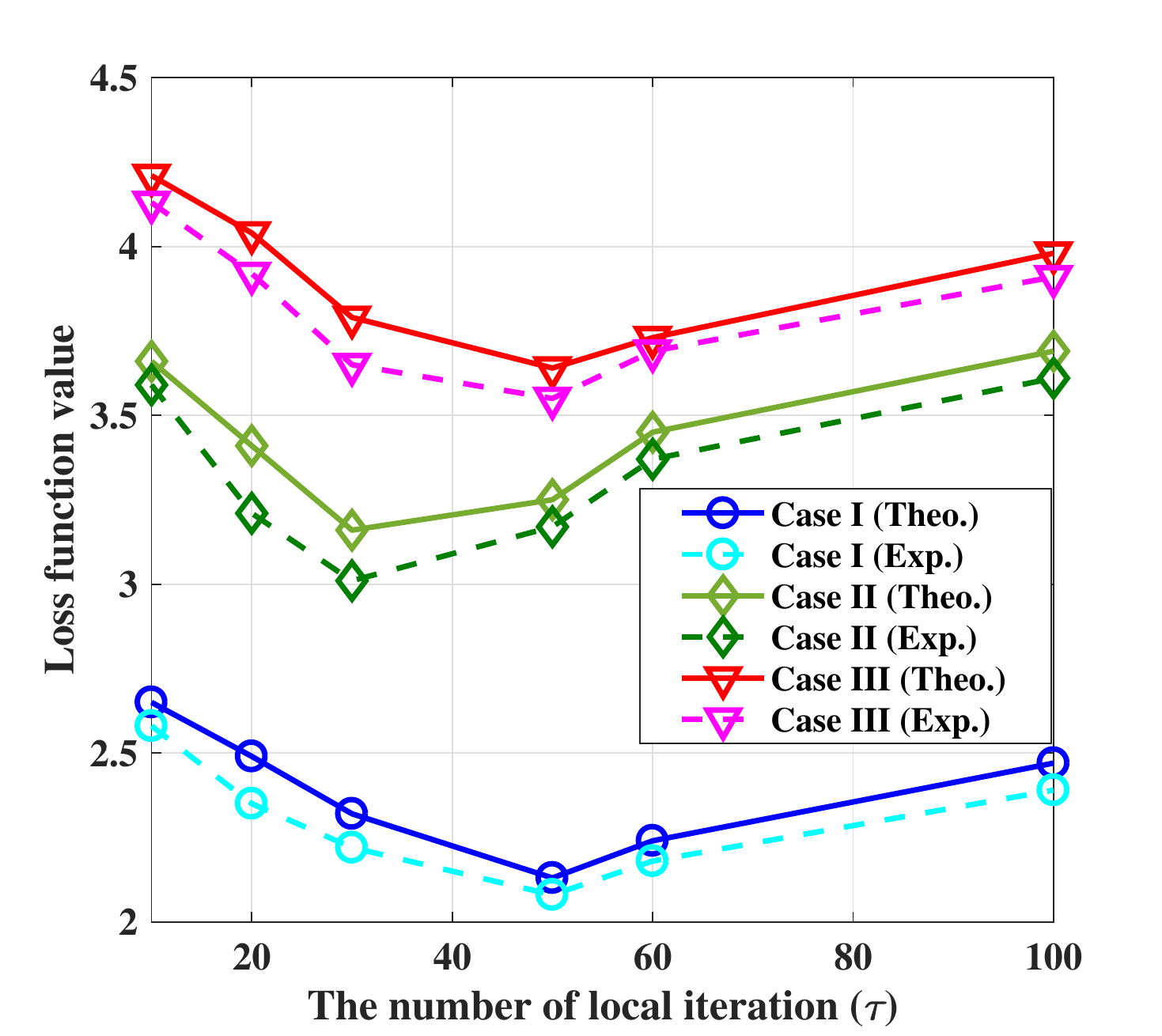}}
\subfigure[Cifar-10 Dataset]{\label{con3}
  \includegraphics[width=0.22\textwidth]{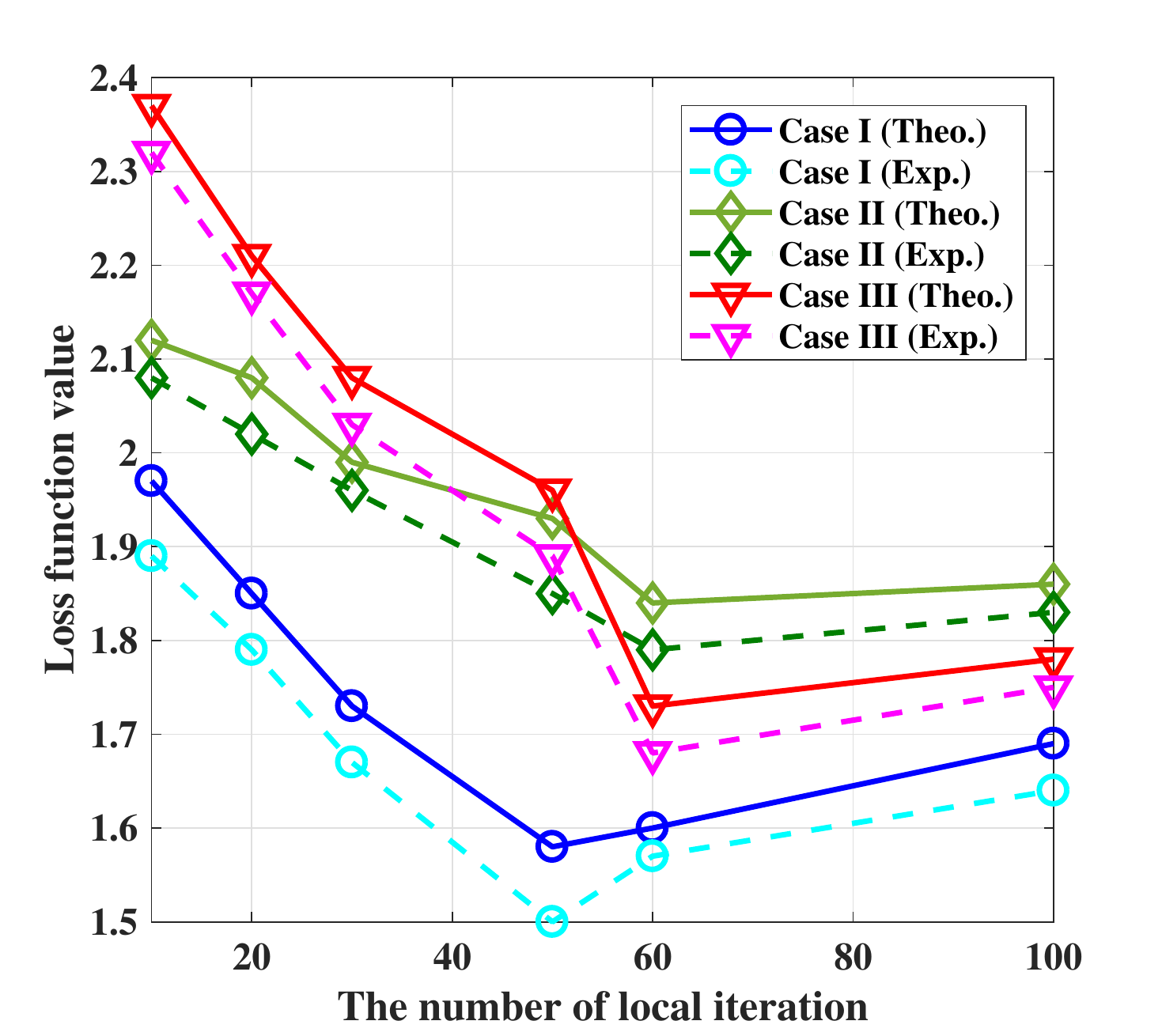}}
\subfigure[Adult Dataset]{\label{con4}
  \includegraphics[width=0.22\textwidth]{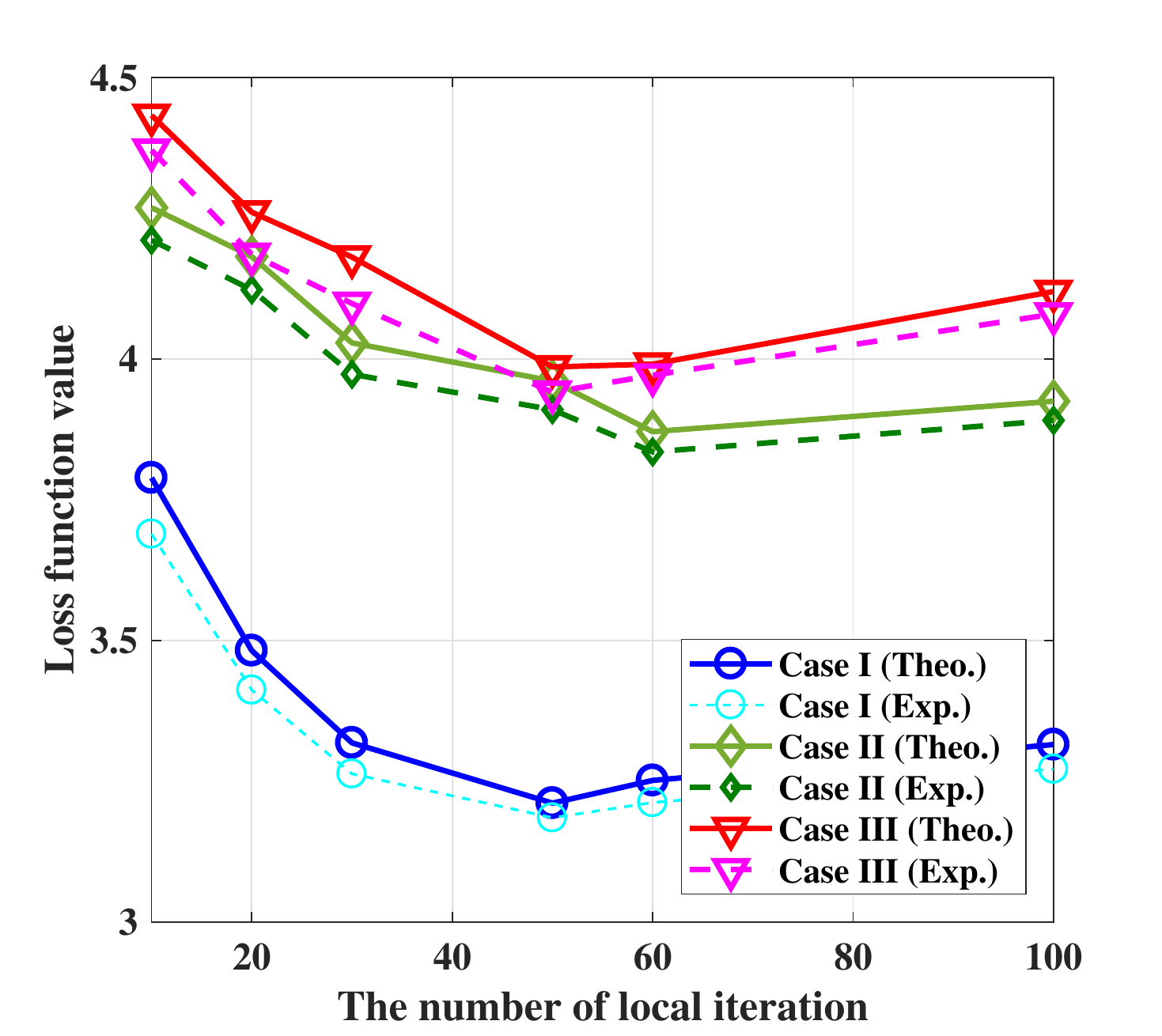}}
 \caption{Comparison of the loss function value between the theoretical and experimental results.} \label{convex}
\end{figure}

In Fig.~\ref{convex}, we show the experimental results of loss function value as a function of $\tau$ with $p_{\textrm{U}} = 0.1$ under the unreliable environment. In order to be close to reality (the local training epoch ($\tau$) and communication rounds ($k$) of clients are not too small) the range of $\tau$ is set to $[10,100]$. We can observe that the theoretical bounds are convex functions and close to the real results for the three cases and four datasets, which are consistent with~\textbf{Theorem~\ref{theorem:bound_reverse}} and~\textbf{Proposition~\ref{proposition:optim_tau}}.
The reason behind this phenomenon is that a large local epochs $\tau$ will decrease the times of uploading unreliable parameters, while a small $\tau$ incurs much unreliability in the parameters uploaded by all clients. {\color{black}In addition, with a smaller value of added noise, the learning performance will get fewer negative influences.}

\subsection{Experimental Results with Unreliable Behavior}
\begin{figure}
\centering
\subfigure[MNIST Dataset]{\label{a1}
  \includegraphics[width=0.22\textwidth]{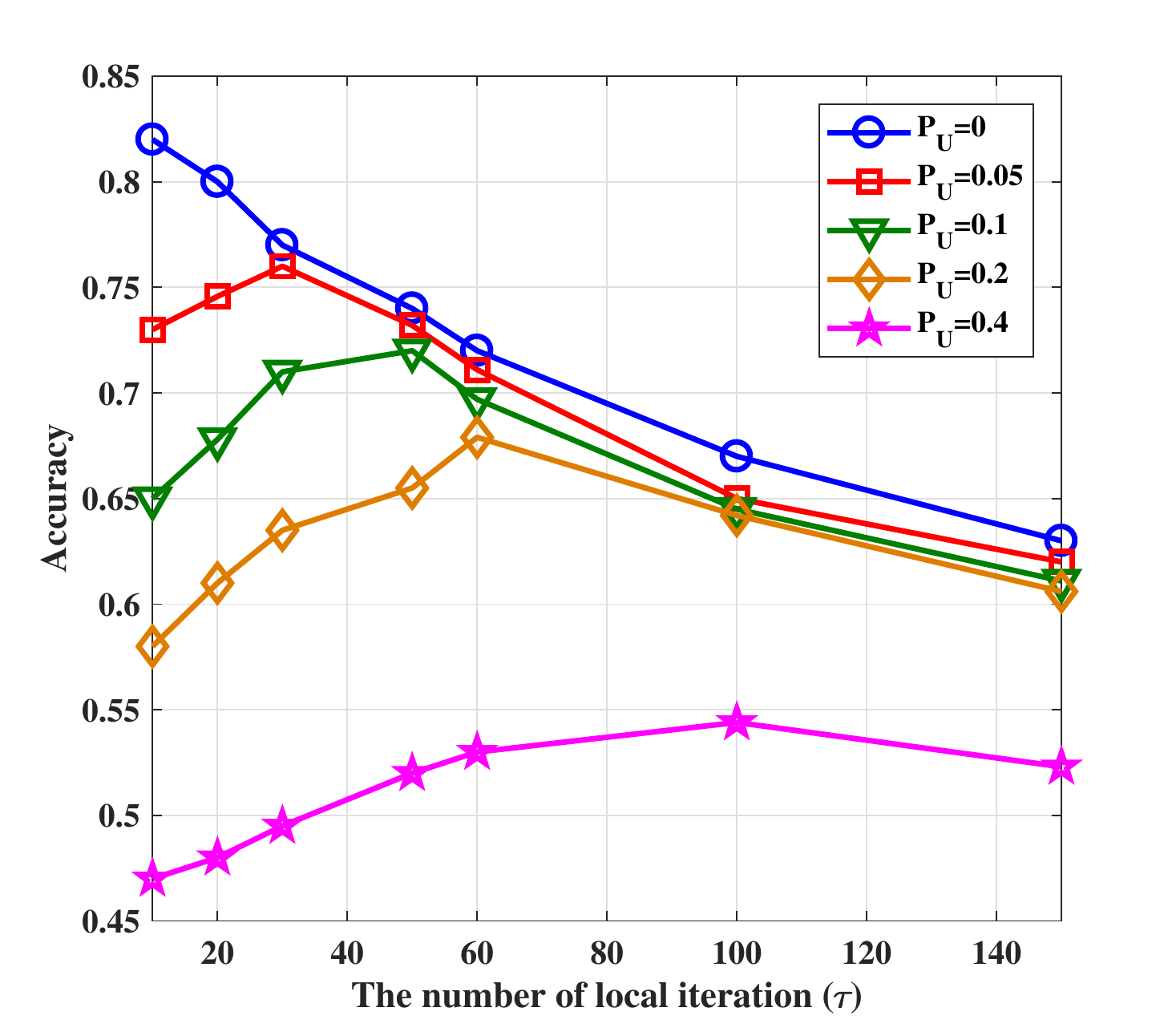}
}
\subfigure[Fashion-MNIST Dataset]{\label{a2}
  \includegraphics[width=0.22\textwidth]{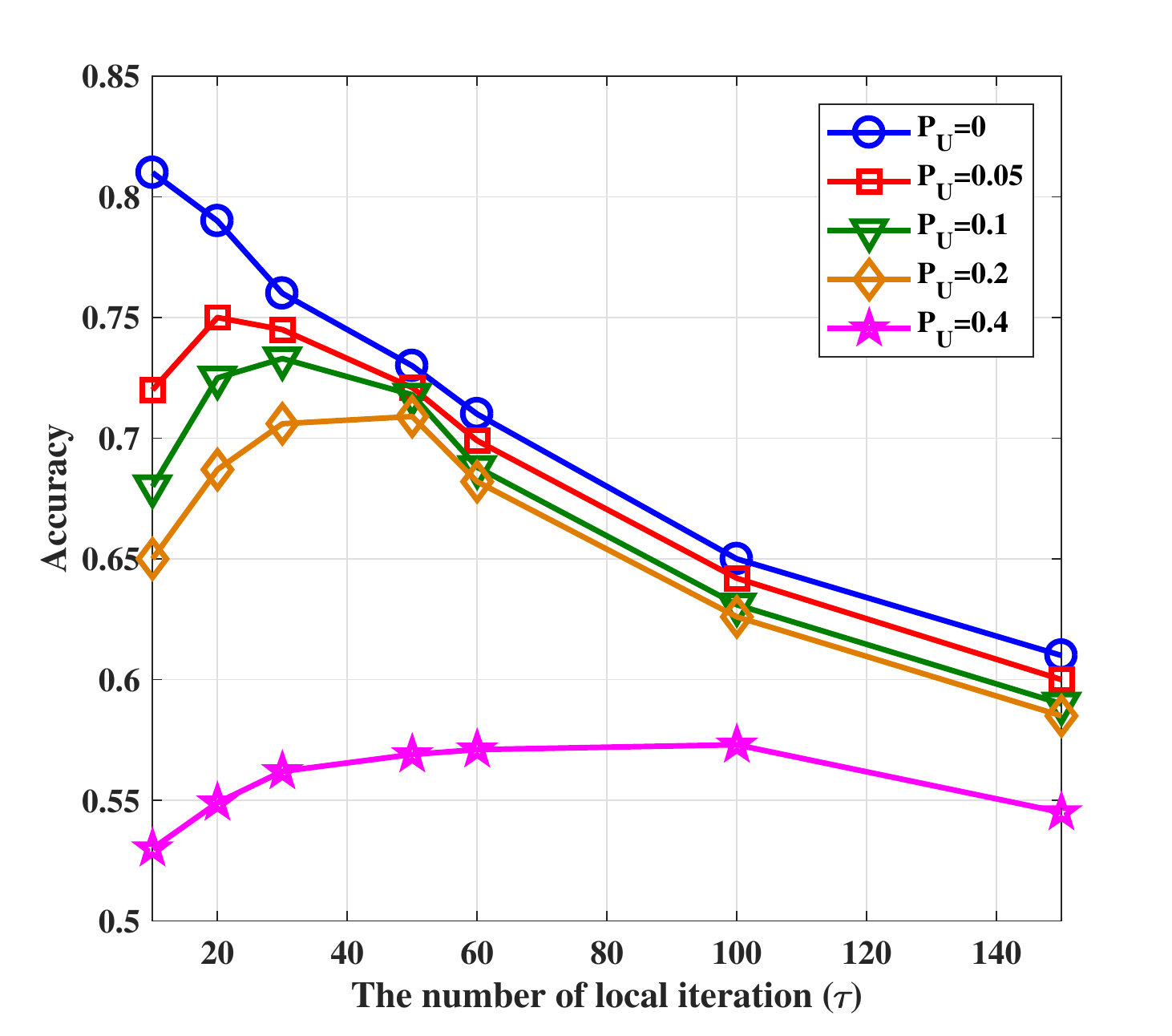}}
  \subfigure[Cifar-10 Dataset]{\label{a3}
  \includegraphics[width=0.22\textwidth]{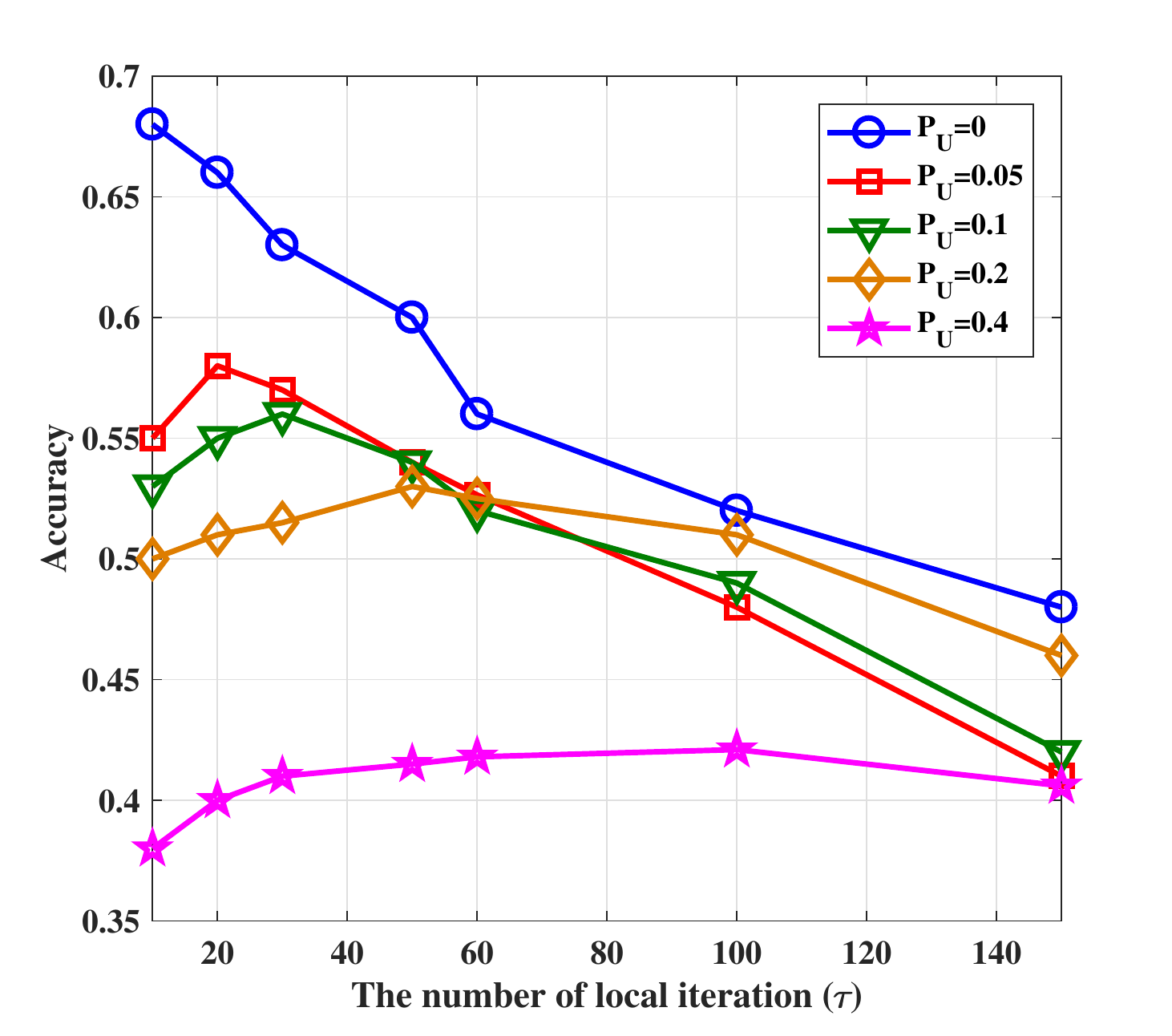}}
  \subfigure[Adult Dataset]{\label{a4}
  \includegraphics[width=0.22\textwidth]{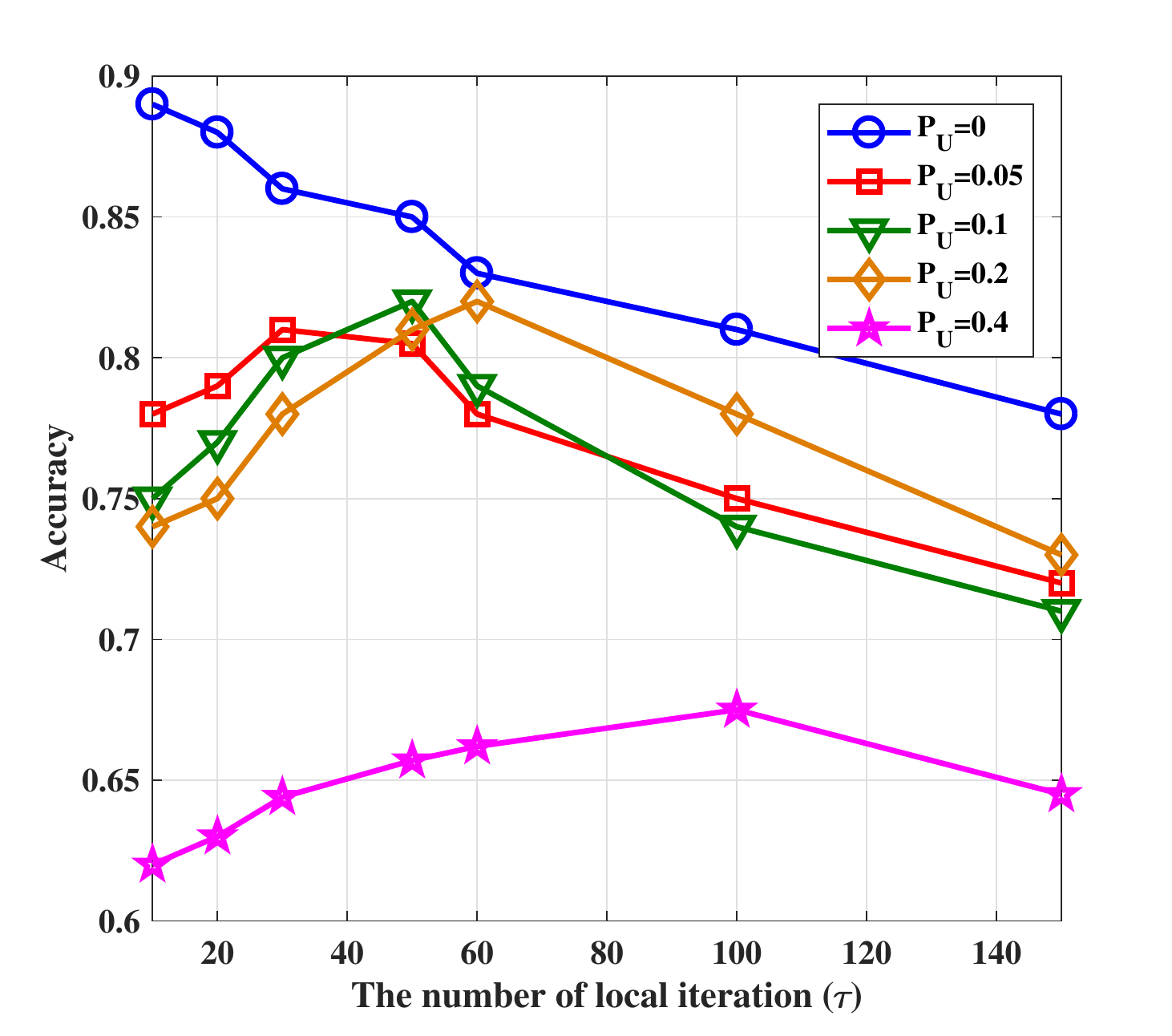}}
 \caption{The classification accuracy of local iterations with a certain probability of unreliable behavior.} \label{acc}
\end{figure}

In this subsection, we show the classification accuracy based on the FL system with different probabilities of unreliable clients in Fig.~\ref{acc}. We take \textbf{Case I} as the abnormal client for MNIST and Fashion-MNIST dataset, \textbf{Case II} for Cifar-10 dataset and \textbf{Case III} for Adult dataset, respectively. In order to show different conditions, we also set the probability of unreliable clients $p_\textrm{U}$ to $0.05$, $0.1$, $0.2$ and $0.4$, respectively. From these figures we find that when there is no abnormal client ($p_\textrm{U}=0$), the system performance decreases with the increase of the local iteration $\tau$, which is consistent with~\textbf{Proposition~\ref{rem:reverse_1}}. However, when the uploading environment is unreliable, we can note that there exits an optimal number of the local iterations $\tau$ in terms of system performance, which is in line with~\textbf{Proposition~\ref{proposition:optim_tau}}.
We can also note that the optimal number of local training iterations increases with the probability of abnormal clients.
The intuition is that more communication rounds will produce a larger damage to the FL system, but more communication rounds also bring a better performance for a normal FL system.
In addition, we find that, with an increasing probability of unreliable probability, the system performance shows a descending trend, and a system with relatively high probability, i.e., $p_{\textrm{U}}>50\%$, may fall to converge.

\begin{figure}
\centering
\subfigure[MNIST Dataset]{\label{m11}
  \includegraphics[width=0.22\textwidth]{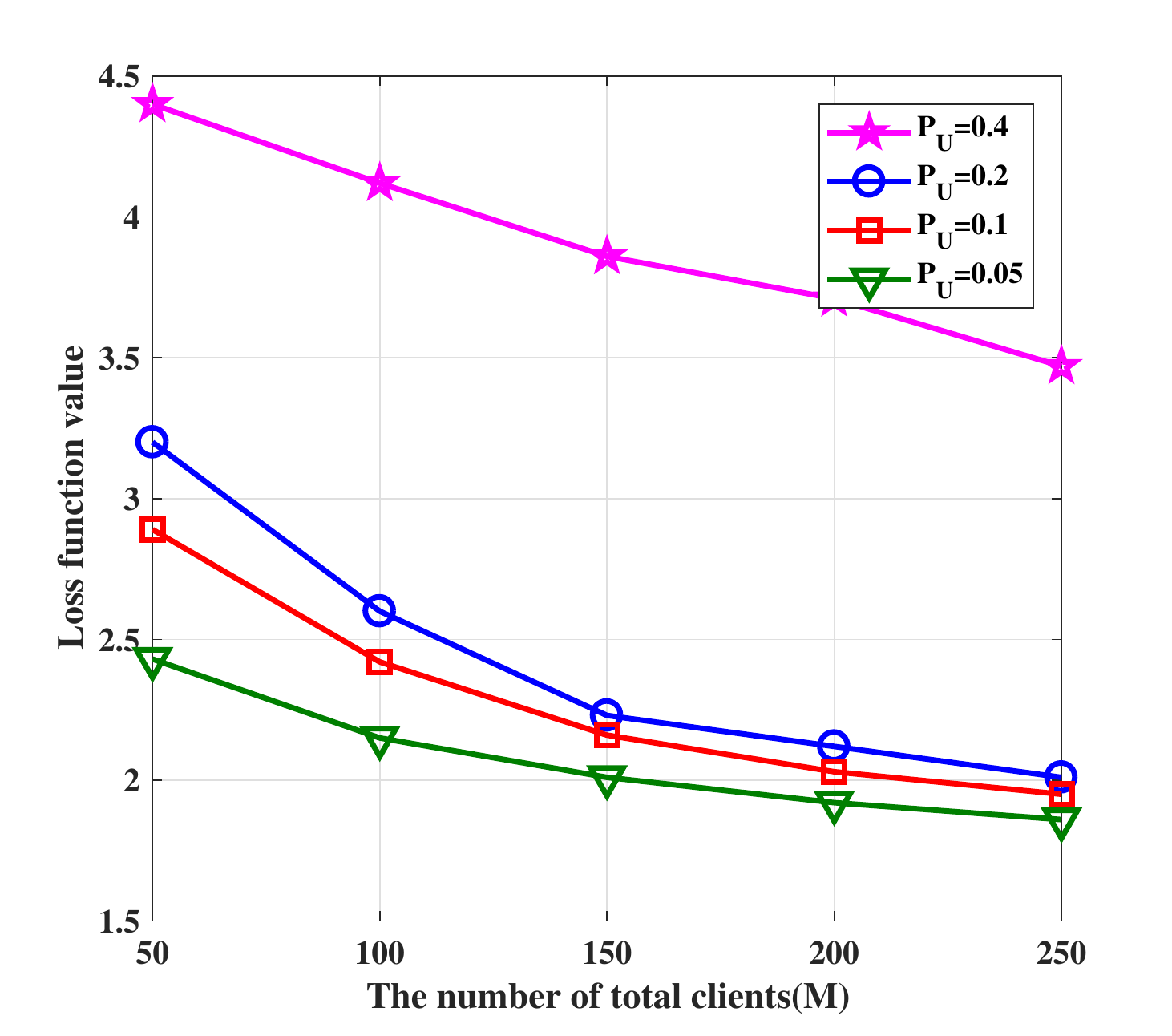}
}
\subfigure[Fashion-MNIST Dataset]{\label{m22}
  \includegraphics[width=0.22\textwidth]{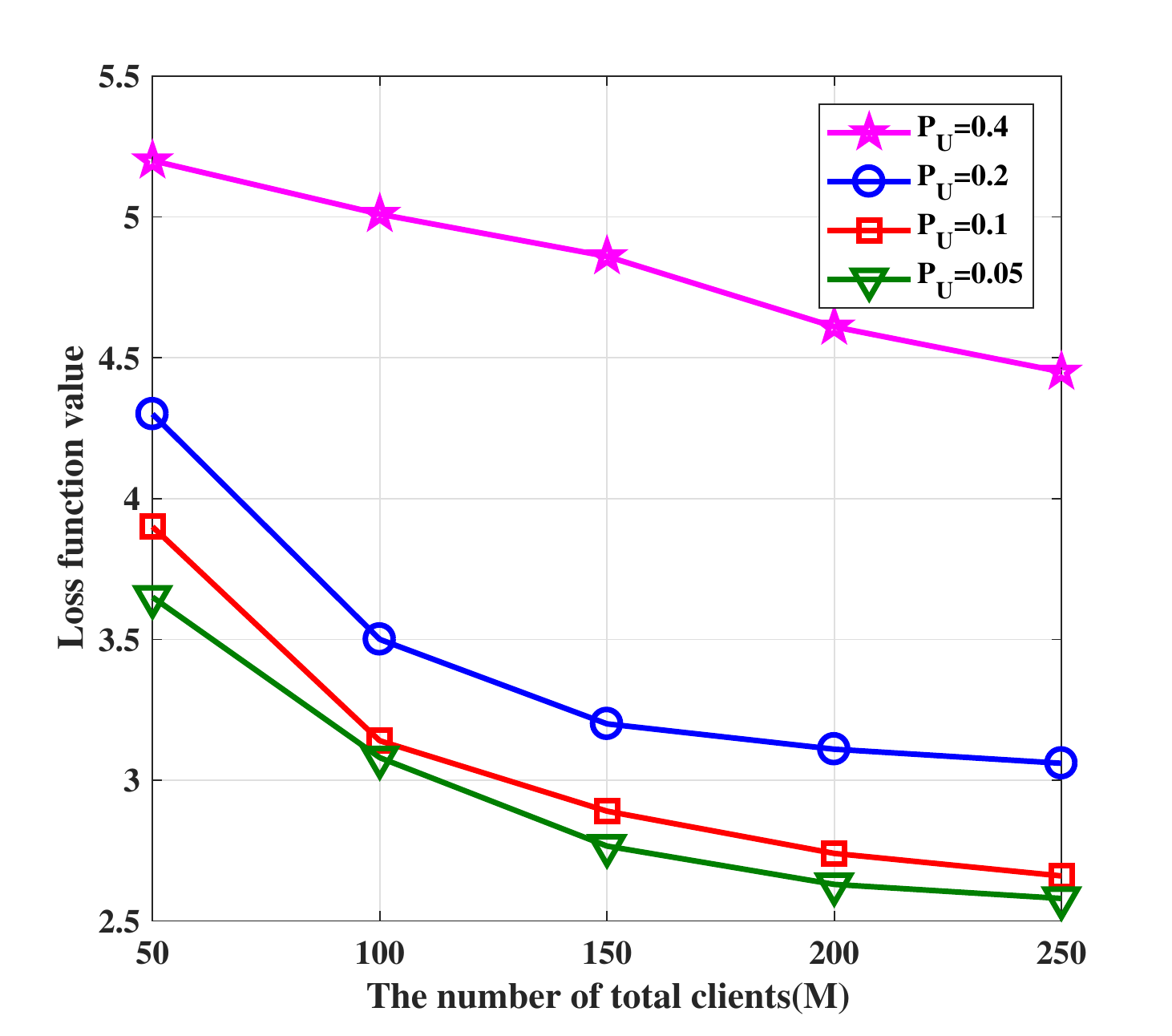}}
 \subfigure[Cifar-10 Dataset]{\label{m33}
  \includegraphics[width=0.22\textwidth]{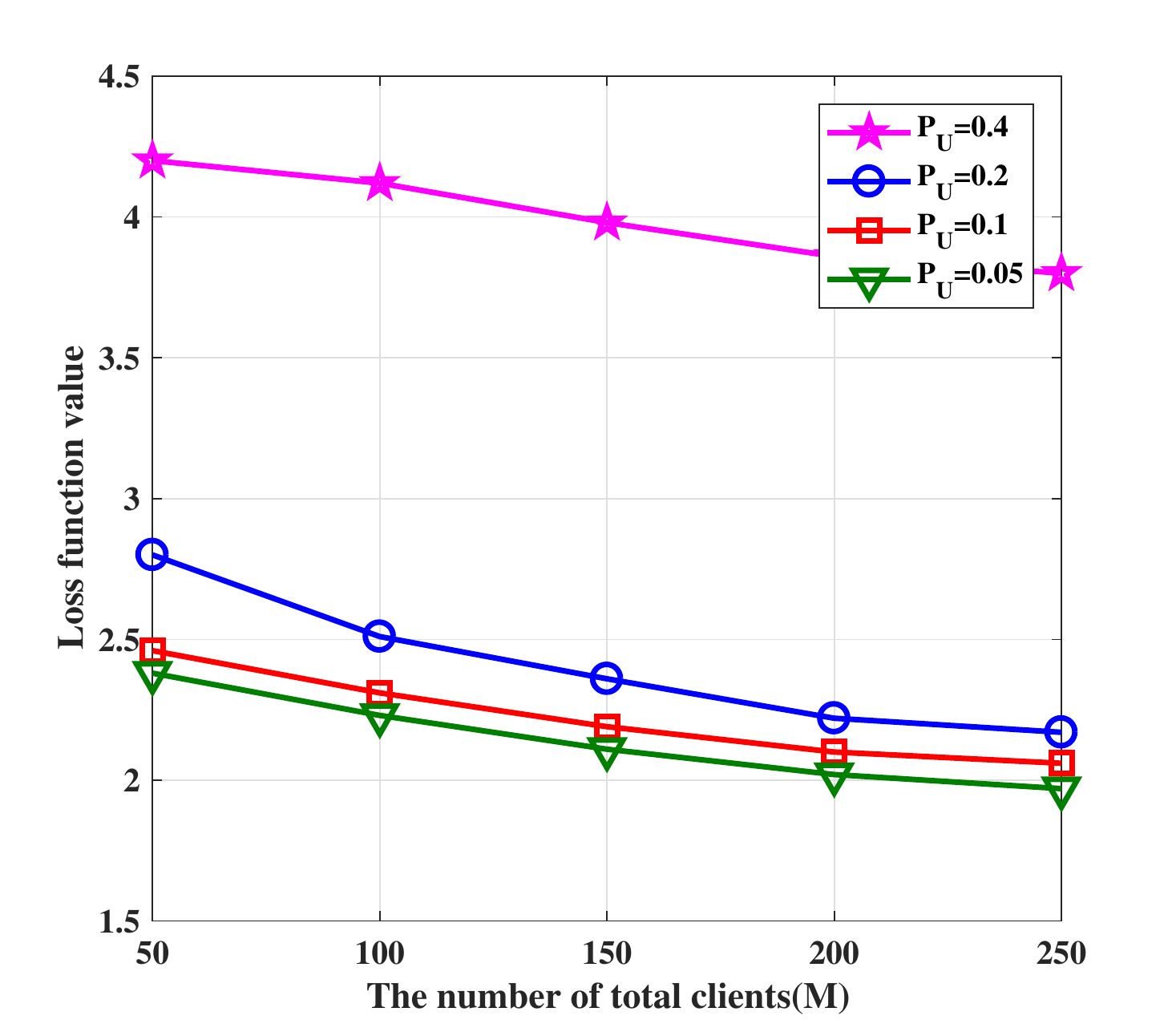}}
  \subfigure[Adult Dataset]{\label{m44}
  \includegraphics[width=0.22\textwidth]{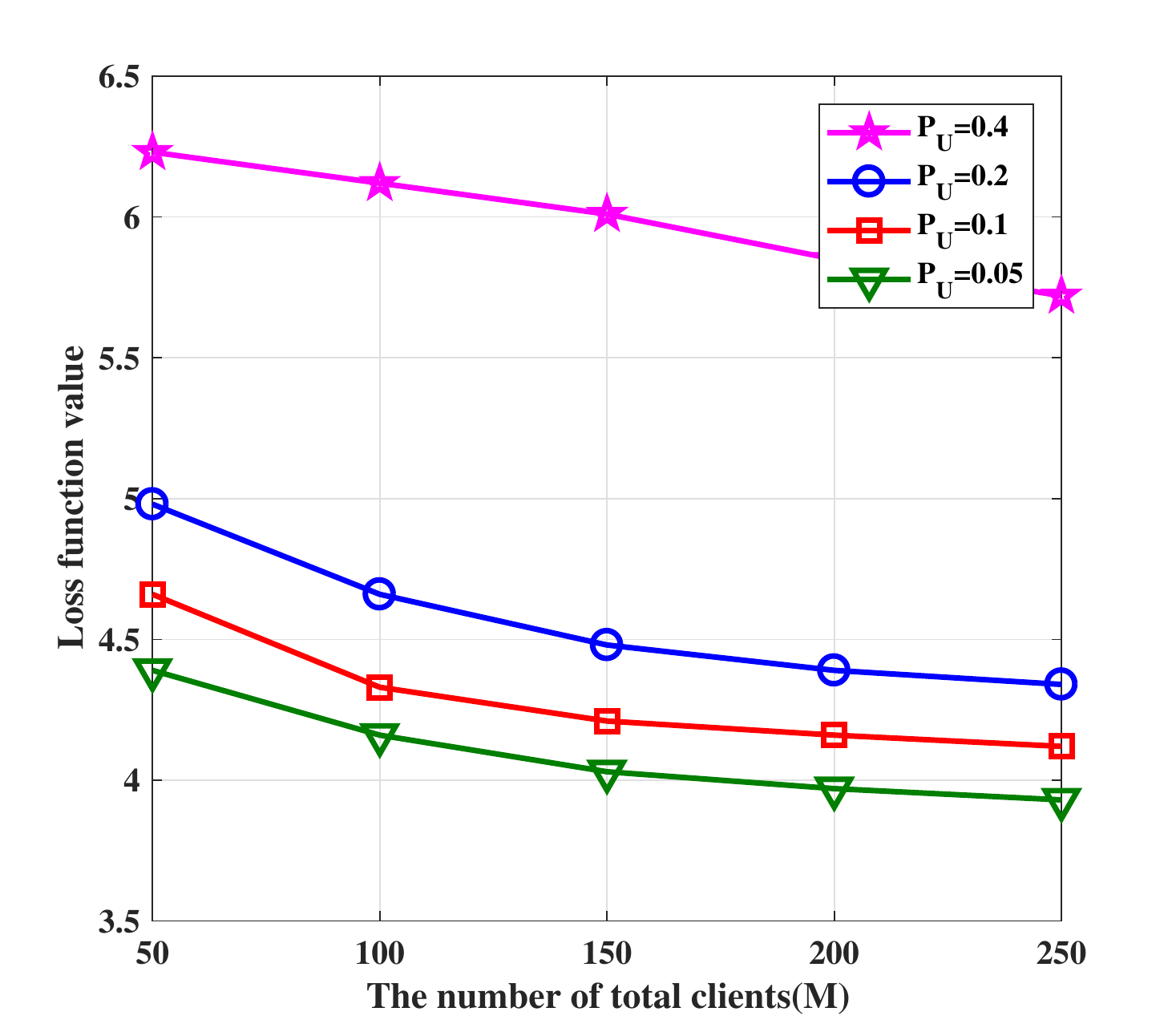}}
 \caption{The loss function value with different number of total clients.} \label{pr_mlp_reverse}
\end{figure}

In Fig.~\ref{pr_mlp_reverse}, we show the loss function value under different numbers of total clients that we set $M$ to $50$, $100$, $150$, $200$ and $250$.
In Fig.~\ref{m11} we use \textbf{Case I}, in Fig.~\ref{m22}-\ref{m33} we use \textbf{Case II}, and in Fig.~\ref{m44} we use \textbf{Case III}, respectively.
We can note that the system has a better performance with a smaller unreliable probability ($p_\textrm{U}$), and the loss function value in both figures keeps decreasing with the number of total clients, which is consistent with~\textbf{Proposition~\ref{cor:reverse_2}}.

\subsection{Performance of the Proposed DeepSA Algorithm}
\begin{figure}[ht]
\centering
\includegraphics[height=2.6in,width=3in,angle=0]{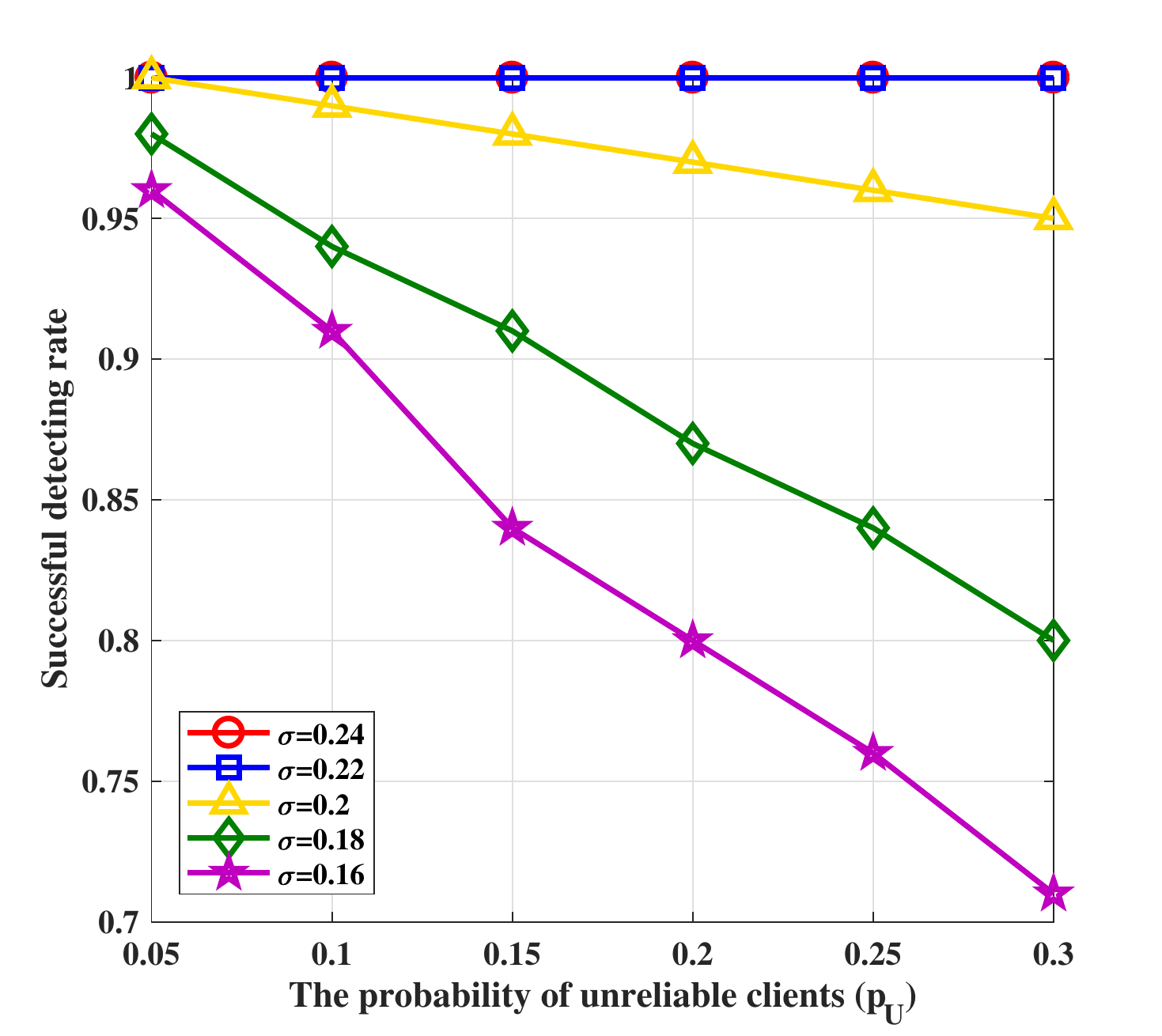}
\caption{The successful detecting rate with a trained DNN detector under a federated model against different probabilities of unreliable clients $p_\textrm{U}$.}
\label{fig:drs_DeepSA}
\end{figure}

In this subsection, we conduct experiments on our proposed DeepSA based federated training against various percentages of unreliable clients.
We use ReLU as the activation functions for the hidden layers: $y =  \text{ReLU}(x) = \max(x, 0)$, where $x \in \mathbb{R}$ is the input, and $y$ is the output of the activation function. To map the output to the interval between $(0,1)$, we choose the sigmoid function as the activation function for the output layer: $y =  \text{sigmoid} (x)= \frac{1}{1+e^{-x}}$.
In Fig.~\ref{fig:drs_DeepSA} we show the detecting results with a stable scalar $\alpha=0.8$, and various noises. From this figure we can observe that with a larger standard deviation $\sigma$, the trained DNN based detector will perform better in identifying these unreliable clients. This is because with a larger standard deviation, a more obvious difference of parameters from neighbouring communication rounds will be recognized by the trained detector. Moreover, it can be noted that when $\sigma > 0.22$, this detector will have an excellent performance, which can guarantee no errors, i.e., the detecting rate is $1$.
We can also find that if $p_{\textrm{U}}$ is larger, the successful detecting rate will decrease which means that it is more difficult for detectors to identify.

\begin{figure}
\centering
\subfigure[MNIST Dataset]{\label{d1}
  \includegraphics[width=0.22\textwidth]{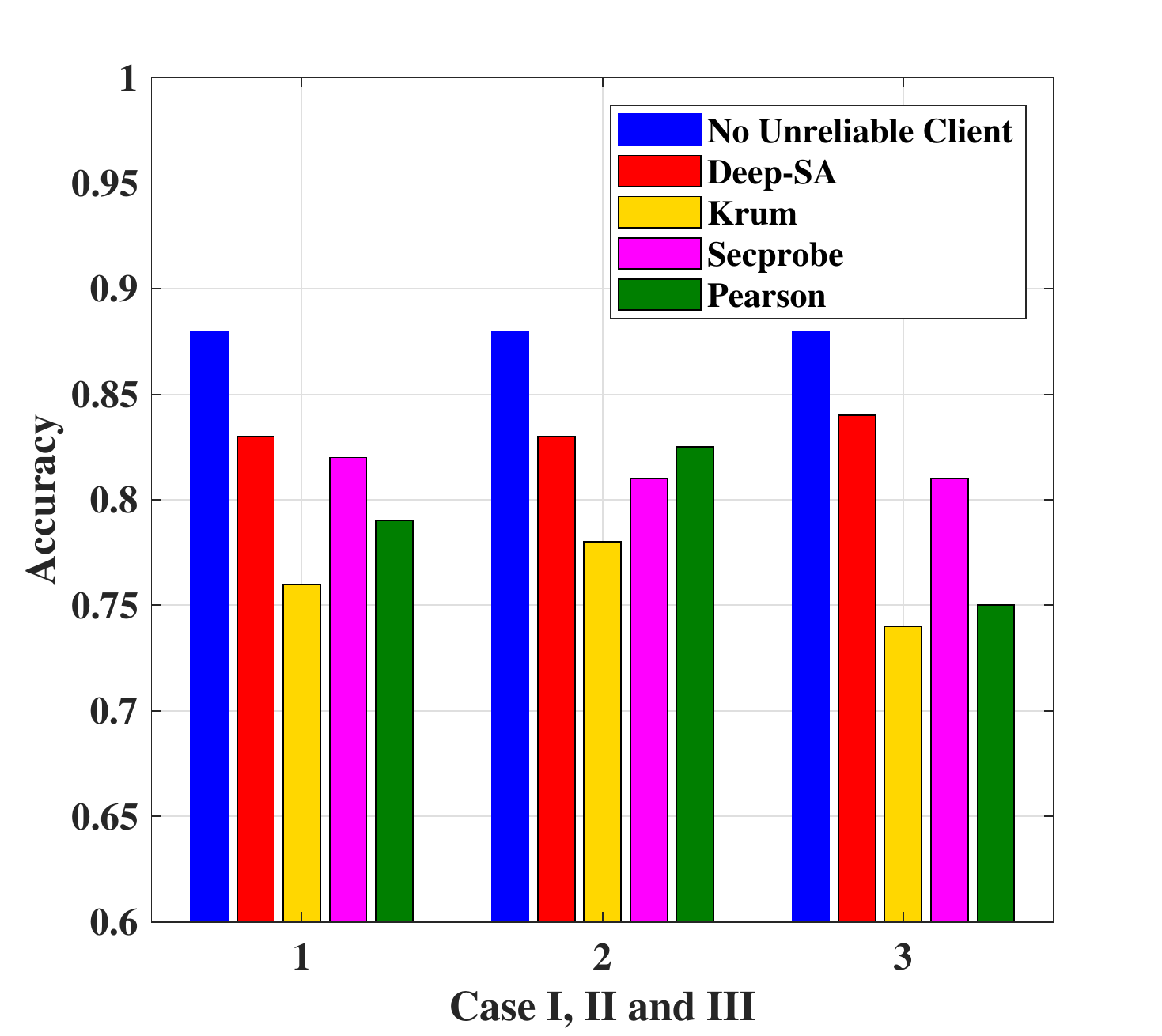}
}
\subfigure[Fashion-MNIST Dataset]{\label{d2}
  \includegraphics[width=0.22\textwidth]{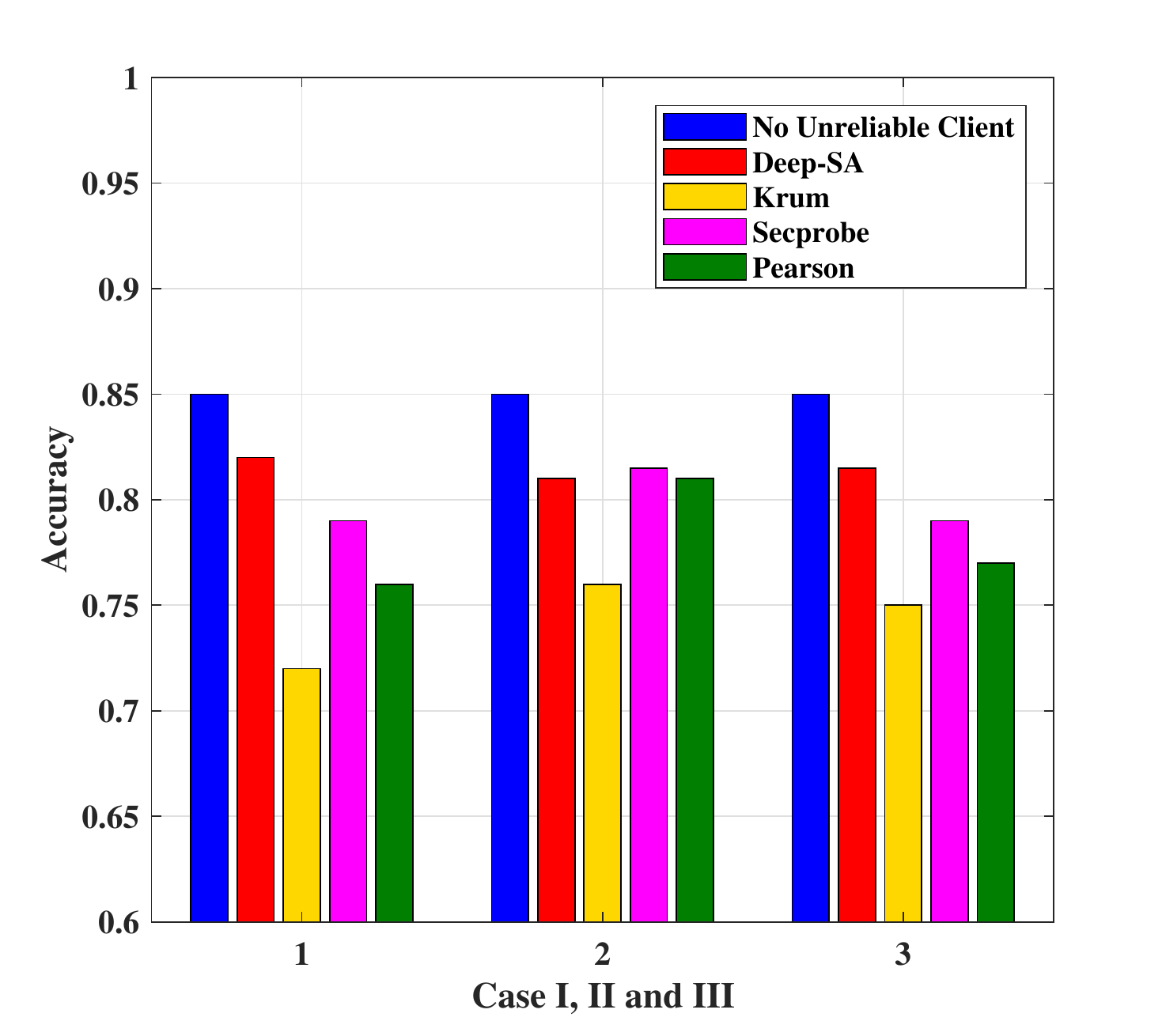}}
 \subfigure[Cifar-10 Dataset]{\label{d3}
  \includegraphics[width=0.22\textwidth]{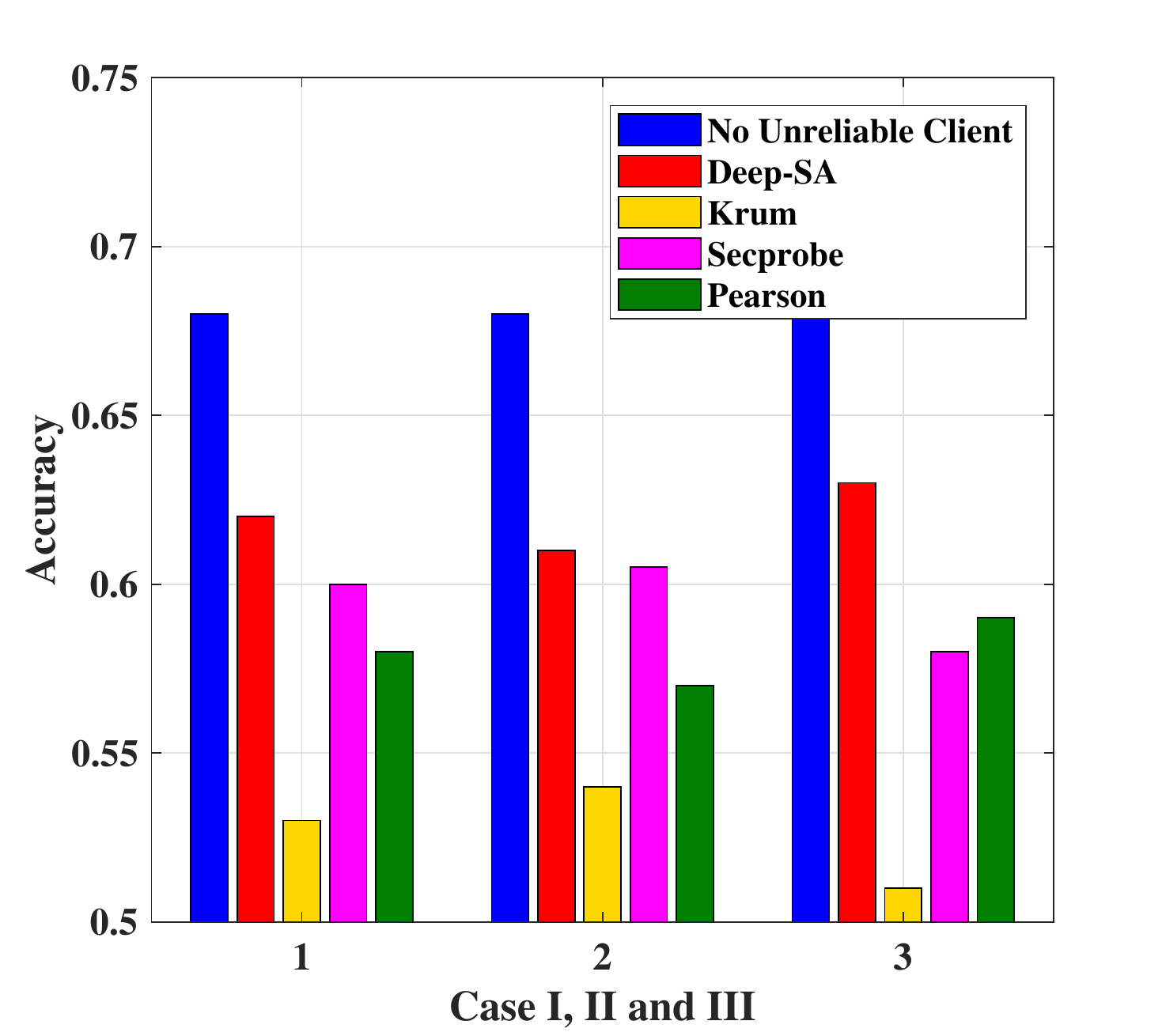}}
  \subfigure[Adult Dataset]{\label{d4}
  \includegraphics[width=0.22\textwidth]{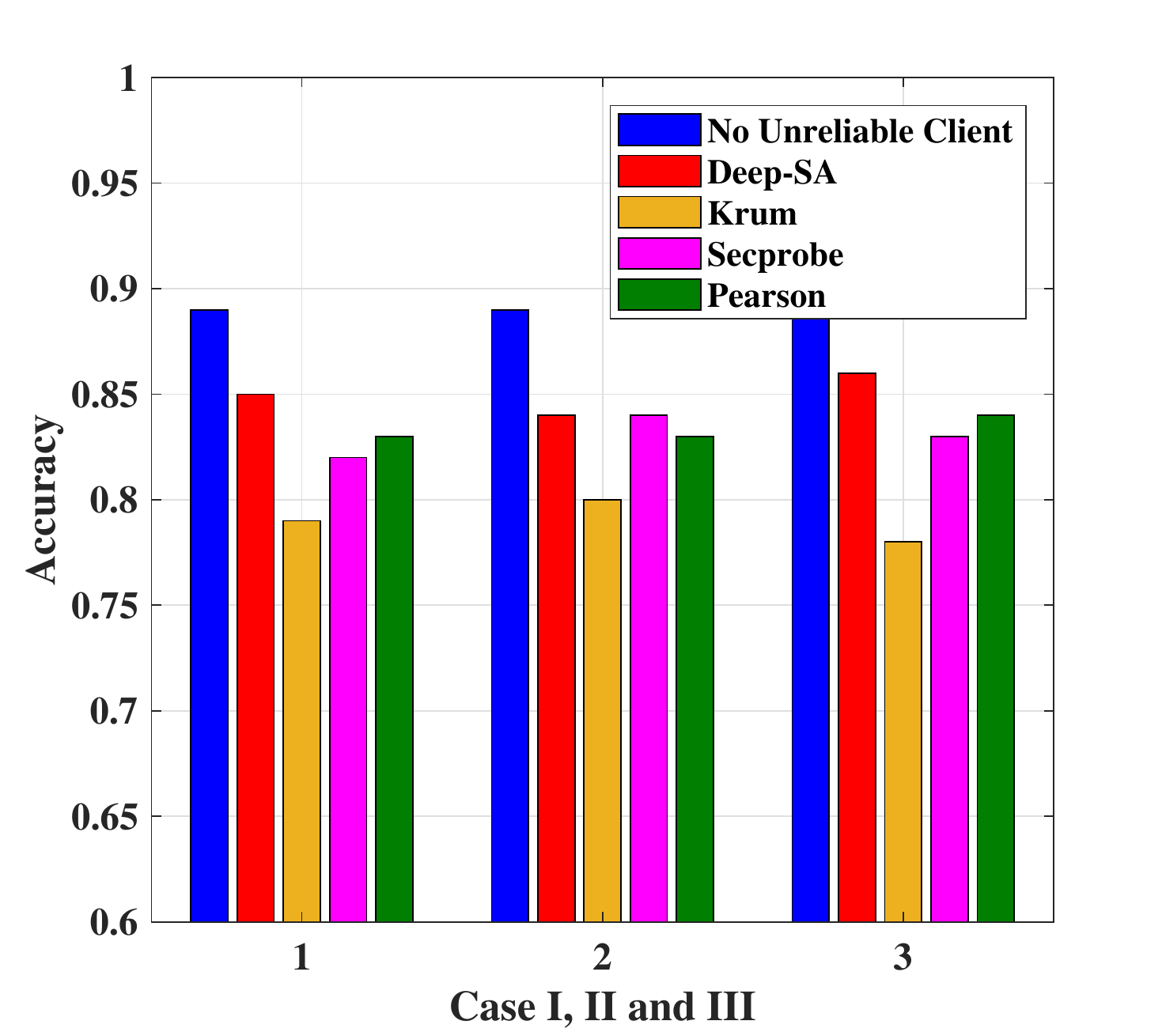}}
 \caption{The classification accuracy comparison between the proposed Deep-SA algorithm and others.} \label{ddd}
\end{figure}

Fig.~\ref{ddd} show the comparison results between the proposed DeepSA algorithm and others, in which we set $p_{\textrm{U}}=0.2$ and unreliable behaviours consider Case I, II and III. In addition, in Table~\ref{t1} and~\ref{t2} we consider a more practical scenario in which clients may behave unreliably with different probabilities. In details, we assume there are 100 clients which are divided into 4 equal-size groups. The probabilities of unreliable behaviours for the 4 groups are set to 0.1, 0.2, 0.3 and 0.4, respectively. It is obvious that with our proposed DeepSA algorithm, the federated training process perform better in most cases. The reason is that the similarity based detection algorithms (Secprobe and Pearson) can only handle the noise perturbation behavior, and Krum loses its performance due to the limited number of participants, while the proposed algorithm has a high detecting rate which enhances the learning performance.

\begin{table}
\caption{The classification accuracy comparison in MNIST/Fashion MNIST dataset with different unreliable probabilities.} \label{t1}
\resizebox{90mm}{10mm}
{
\begin{tabular}{|c|c|c|c|c|c|}
  \hline
  {} & All reliable & Deep-SA & Krum & Secprobe & Pearson \\
  \hline
  Case I & 0.88/0.85 & \textbf{0.81}/\textbf{0.78} & 0.71/0.69 & 0.78/0.75 & 0.74/0.71 \\
  \hline
  Case II & 0.88/0.85 & \textbf{0.8}/\textbf{0.765} & 0.72/0.69 & 0.77/0.74 & 0.76/0.72 \\
  \hline
  Case III & 0.88/0.85 & \textbf{0.81}/\textbf{0.77} & 0.73/0.7 & 0.78/0.75 & 0.74/0.7  \\
  \hline
\end{tabular}
}
\end{table}

\begin{table}
\caption{The classification accuracy comparison in Cifar-10/Adult dataset with different unreliable probabilities.} \label{t2}
\resizebox{90mm}{10mm}
{
\begin{tabular}{|c|c|c|c|c|c|}
  \hline
  {} & All reliable & Deep-SA & Krum & Secprobe & Pearson \\
  \hline
  Case I & 0.68/0.89 & \textbf{0.58}/\textbf{0.82} & 0.51/0.77 & 0.57/0.8 & 0.54/0.78 \\
  \hline
  Case II & 0.68/0.89 & \textbf{0.59}/\textbf{0.83} & 0.52/0.79 & 0.57/0.81 & 0.56/0.79 \\
  \hline
  Case III & 0.68/0.89 & \textbf{0.6}/\textbf{0.83} & 0.52/0.78 & 0.58/0.81 & 0.54/0.78 \\
  \hline
\end{tabular}
}
\end{table}

In addition,  the proposed defensive mechanism is applied to four real-world datasets, i.e., Sports\footnote{https://archive.ics.uci.edu/ml/datasets/Daily+and+Sports+Activities.}, UAV Detection\footnote{https://archive.ics.uci.edu/ml/datasets/Unmanned+Aerial+Vehicle+\\\%28UAV\%29+Intrusion+Detection.}, Energy\footnote{https://archive.ics.uci.edu/ml/datasets/Energy+efficiency.} and Space Shuttle\footnote{https://archive.ics.uci.edu/ml/datasets/Statlog+(Shuttle).}, which have been collected from real-life sensors \cite{9146846,8422402}, and the descriptions of these datasets are listed as follows:
\begin{itemize}
  \item Sports: This dataset comprises motion sensor data of 19 daily and sports activities each performed by 8 subjects in their own style for 5 minutes, and we evaluate the performance by the accuracy of a 19-class classifier.
  \item UAV Detection: This dataset consists of 55 attributes in which each data row represents an encrypted WiFi traffic record. The output shows the current traffic is from a UAV or not.
  \item Energy: This dataset consists of assessing the heating load and cooling load requirements of buildings (that is, energy efficiency) as a function of building parameters with 8 attributes, and aims to predict each of two responses.
  \item Space Shuttle: This shuttle dataset contains 9 attributes and 58000 numerical instances with an 80\% default accuracy.
\end{itemize}
\begin{figure}
\centering
\subfigure[Sports Dataset]{\label{iot1}
  \includegraphics[width=0.22\textwidth]{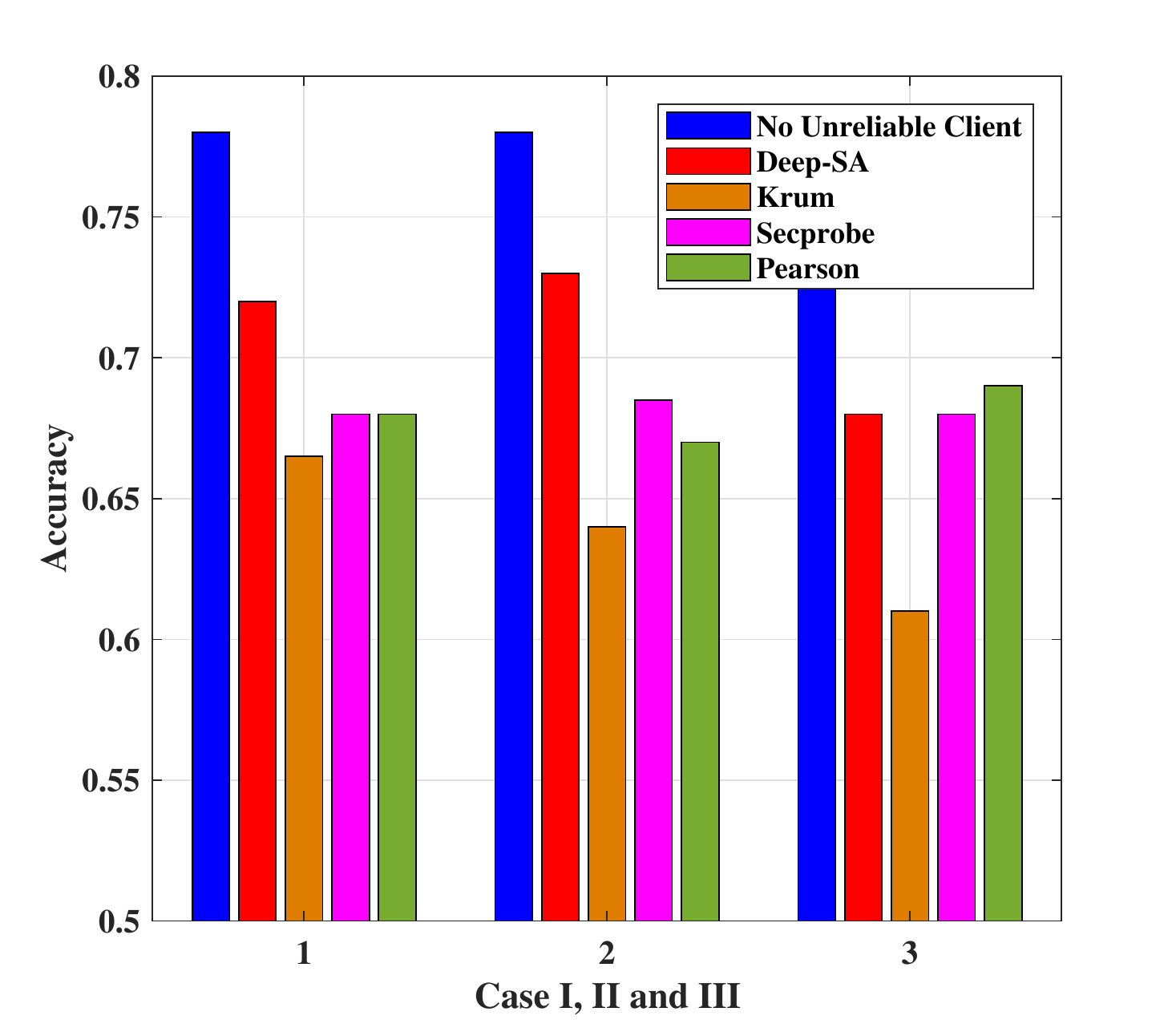}
}
\subfigure[UAV Detection Dataset]{\label{iot2}
  \includegraphics[width=0.22\textwidth]{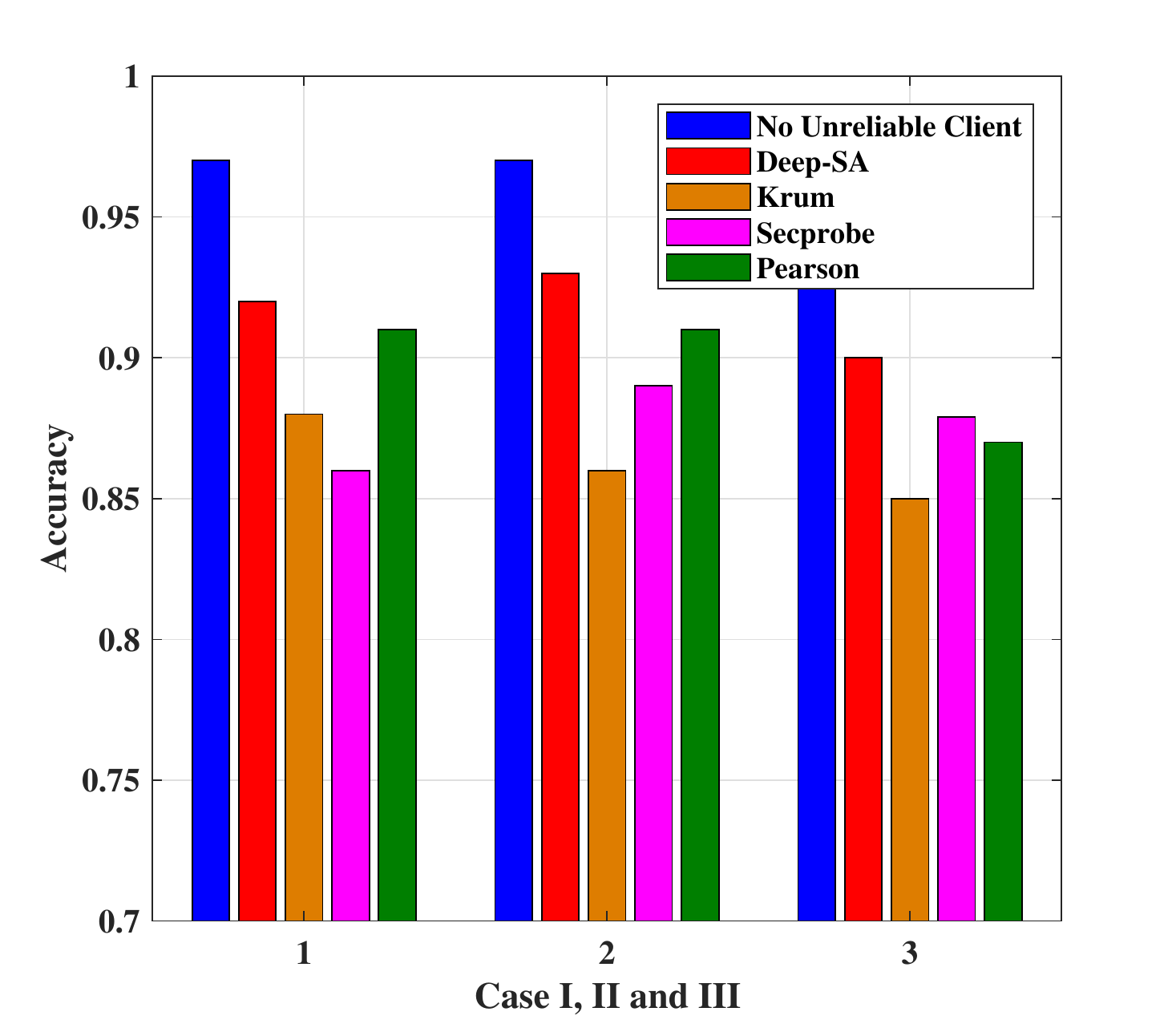}}
 \subfigure[Energy Dataset]{\label{iot3}
  \includegraphics[width=0.22\textwidth]{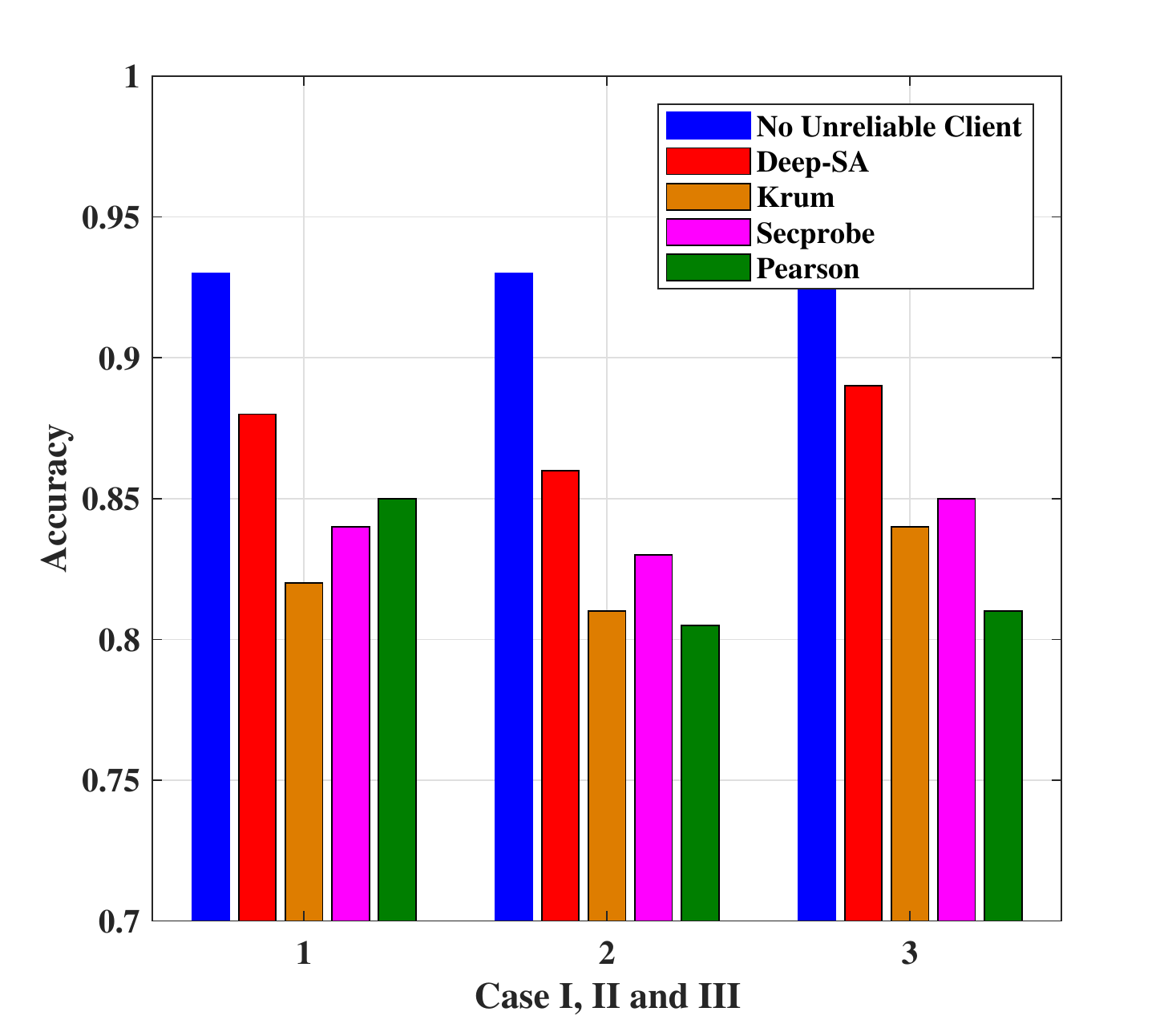}}
  \subfigure[Space Shuttle Dataset]{\label{iot4}
  \includegraphics[width=0.22\textwidth]{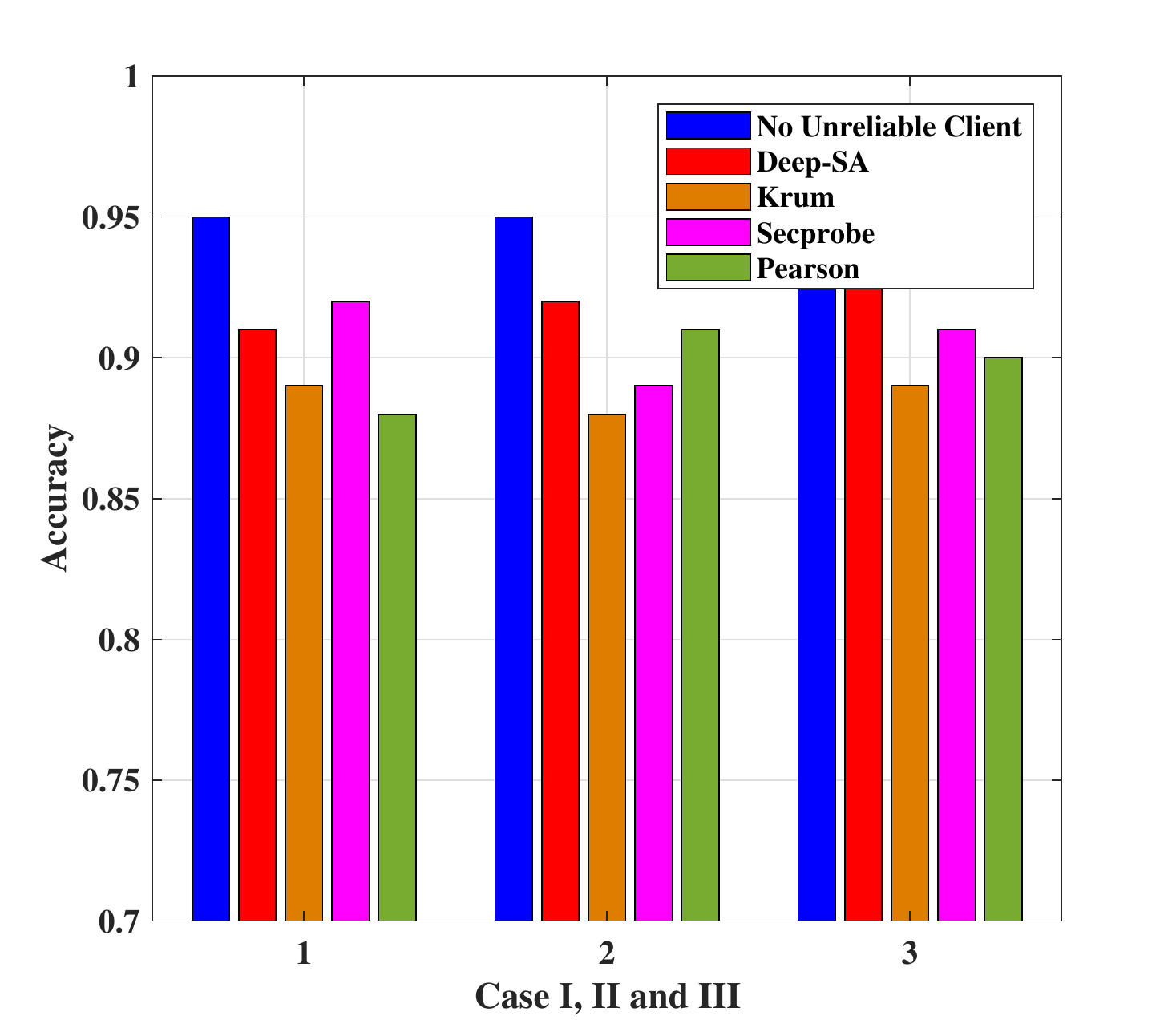}}
 \caption{The classification accuracy comparison between the proposed Deep-SA algorithm and others.} \label{iot}
\end{figure}
In Fig.~\ref{iot}, we verify the proposed defensive algorithm under Case I, II and III of unreliable clients. As can be found in this figure, although some algorithms may have a slightly better or equal performance than the proposed algorithm, Deep-SA outperforms other algorithms in most cases. For example, in the Sports dataset, although DeepSA has not achieved the top performance in Case III (68\% v.s. 69\%), it has a performance gain in other cases (4\% in Case I and 4.5\% in Case II, respectively).

\section{Related Works}
In this section, we investigate different adversarial models in FL and defensive mechanisms, which are active areas of research.
\subsection{Adversarial Models in Federated Learning}
The security of machine learning has attracted heated attentions recently \cite{biggio2012poisoning,biggio2013security}. Although the data is not explicitly exposed in the original format in distributed learning frameworks, e.g., FL \cite{ma2020safeguarding}, different types of adversarial models against distributed machine learning algorithms have been designed and analyzed including poisoning attacks (e.g., \cite{fang2019local,bhagoji2019analyzing}) and privacy attacks (e.g., \cite{hitaj2017deep,wang2019beyond,melis2019exploiting}). For example, poisoning attackers can control part of clients and manipulate the outputs sent to the server, which can mislead the global model deviate to the designed direction \cite{fang2019local}. The authors in \cite{baruch2019little} proposed a novel attacking method that a malicious opponent may interfere with the learning process by applying limited changes to the uploaded models. In addition, the authors in \cite{bhagoji2019analyzing} explored the adversarial of model poisoning attacks on FL, which supported by a single, non-colluding malicious client where the adversarial objective is to make the global model misclassify a set of chosen inputs with high confidence.
\subsection{Defensive Mechanisms}
With the development of adversarial models in FL, how to design an effective defensive mechanism to defeat these malicious clients has became crucial. For detecting poisoned updates in the collaborative learning \cite{9066920}, the results of client-side cross-validation were applied for adjusting the weights of the updates when performing aggregation, where each update is evaluated over other clients' local data. Similar approach based on Pearson similarity is proposed in \cite{7783532}. In addition, the authors \cite{zhao2019privacy,8825829} in considered the existence of unreliable clients in Fl and used the auxiliary validation data to compute a utility score for each participant, thus reducing the negative impact of these unreliable participants. The work in \cite{NIPS2017_f4b9ec30} proposed a novel poisoning defensive method in Fl. In details, for each client, the server will calculate the sum of the Euclidean distances to the models of other clients, and select the one with the minimum sum. However, the mentioned defensive algorithms all need an online detection process while the access to the auxiliary dataset may leak privacy.
\section{Conclusion}\label{sec:concl}
In this work, we have introduced a new threat model of adversary clients for FL systems. By deriving convergence bound on the loss function of the trained FL model, we have seen that there exists an optimal number of local training iterations to achieve the best performance with a fixed total amount of computing resources. Furthermore, we have designed a novel defensive algorithm using the DNN detection technique, termed DeepSA, which can automatically detect unreliable models and remove them from the aggregation process. Extensive experimental results have validated our analysis and the effectiveness of the proposed DeepSA algorithm.
\appendices
\section{Proof of Theorem~\ref{theorem:bound_reverse}}\label{appendix:bound_reverse}
Taking the expectation of the unreliable behavior, we transform  Eq.~\eqref{equ:reversed_model} as
\begin{equation}
\begin{aligned}
\bar{\boldsymbol{w}}^{(k\tau)}&=\frac{1}{M}\sum\limits_{i \in M} \left[ (1 - {p_\textrm{U}})\boldsymbol{w}_{i} + {p_\textrm{U}}\hat {\boldsymbol{w}_i} \right].
\end{aligned}
\end{equation}
Then, according to~\textbf{Assumption 1}, the difference between $F(\bar{\boldsymbol{w}}(k\tau))$ and $F(\boldsymbol{w}(k\tau))$ can be expressed as
\begin{equation}\label{faa}
\begin{aligned}
&F(\bar{\boldsymbol{w}}^{(k\tau)})- F(\boldsymbol{w}^{(k\tau)})\leq\rho\left\Vert\bar{\boldsymbol{w}}^{(k\tau)}-\boldsymbol{w}^{(k\tau)}\right\Vert\\
&=\frac{\rho }{M}\left\Vert {\sum\limits_{i \in M} {{p_\textrm{U}}\left( {\hat{\boldsymbol{w}}_i - \boldsymbol{w}_i} \right)} } \right\Vert \\
&=\frac{\rho }{M}\left\Vert {\sum\limits_{i \in M} {{p_\textrm{U}}\left[ {(\alpha  - 1)\boldsymbol{w}_i + \boldsymbol{n}_i} \right]} } \right\Vert\\
&\leq \frac{{\rho {p_\textrm{U}}}}{M}\left[\mathbb{E}\left\Vert\sum\limits_{i \in M} {(\alpha   - 1)\boldsymbol{w}_i} \right\Vert + \mathbb{E}\left\Vert\sum\limits_{i \in M} \boldsymbol{n}_i \right\Vert\right]\\
&\leq \frac{{\rho {p_\textrm{U}}}}{M}\left[ {(1 - \alpha  )M\Theta  + \frac{{2\sqrt M \sigma }}{\pi }} \right],
\end{aligned}
\end{equation}
where $\Vert \Vert$ denotes the $L_2$ norm function, $\Theta$ is a upper bound of all model parameters, respectively. For simplicity, we omit the superscript ($k\tau$) of $\boldsymbol{w}_i$ and $\boldsymbol{n}_i$.

\begin{figure}[ht]
\centering
\includegraphics[height=2.4in,width=3.2in,angle=0]{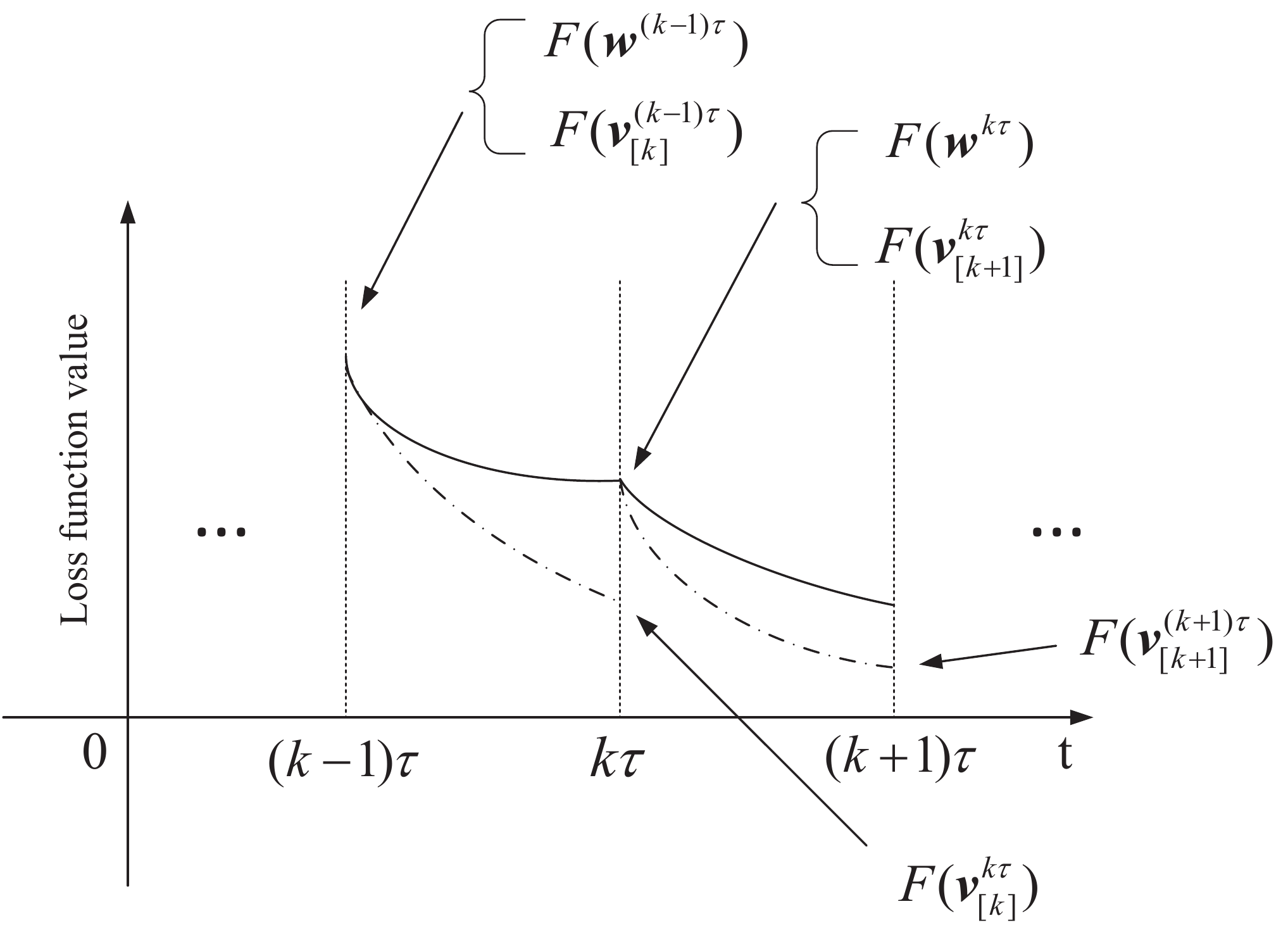}
\caption{The auxiliary parameter vector in FL.}
\label{auxiliary}
\end{figure}

We define $\boldsymbol{v}_{[k]}^{(t)}$ as an auxiliary parameter vector, which follows an centralized gradient decent.
\begin{equation}
\boldsymbol{v}_{[k]}^{(t)} = \boldsymbol{v}_{[k]}^{(t-1)} - \eta\nabla F(\boldsymbol{v}_{[k]}^{(t-1)}).
\end{equation}
The notation $[k]$ defines the interval $[(k-1)\tau, k\tau]$ for $k = 1,2,\ldots,K$, and the auxiliary parameter vector $\boldsymbol{v}_{[k]}^{(t)}$ only works in this interval. We denote by $\boldsymbol{v}_{[k]}^{(k\tau)}$ and $\boldsymbol{v}_{[k+1]}^{(k\tau)}$ the two different auxiliary vectors before and after the $k$th aggregation respectively, which can be seen in Fig.~\ref{auxiliary}. At the beginning of the interval $[k]$, $\boldsymbol{v}_{[k]}^{(t)}$ will not inherit the last result of the previous interval but is equivalent to the global parameter after $k$th aggregation. i.e., $\boldsymbol{v}_{[k+1]}^{(k\tau)} \triangleq \boldsymbol{w}^{(k\tau)}$.

Using the upper bound in~\cite{Wang2019Adaptive}, we know that
\begin{equation}\label{gdg}
F(\boldsymbol{w}^{(k\tau)})-F(\boldsymbol{v}_{[k]}^{(k\tau)})\leq\rho \phi(\tau),
\end{equation}
where $k$ is the index of the aggregation, and $\phi(\tau)=\frac{\delta}{\beta}\left((\eta\beta+1)^{\tau}-1\right)-\eta\delta \tau$.

Combing Eq.~(\ref{faa}) with Eq.~(\ref{gdg}) we can obtain
\begin{equation}
\begin{aligned}
&F(\bar{\boldsymbol{w}}^{(k\tau)})-F(\boldsymbol{v}_{[k]}^{(k\tau)})\leq \rho \phi(\tau)+\frac{{\rho {p_\textrm{U}}}}{M}\left[ {(1 - \alpha )M\Theta  + \frac{{2\sqrt M \sigma }}{\pi }} \right]\\
&=\rho\left\{\phi(\tau)+\frac{{ {p_\textrm{U}}}}{M}\left[ {(1 - \alpha  )M\Theta  + \frac{{2\sqrt M \sigma }}{\pi }} \right]\right\}=\rho\Delta,
\end{aligned}
\end{equation}
where $\phi(\tau)+\frac{{ {p_\textrm{U}}}}{M}\left[ {(1 - \alpha  )M\Theta  + \frac{{2\sqrt M \sigma }}{\pi }} \right]$ is denoted by $\Delta$.
Then, we define $\theta_{[k]}^{(t)}=F\left(\boldsymbol{v}_{[k]}^{(t)}\right)-F(\boldsymbol{w}^{*})$ for an interval $[k]$ where $k$ is fixed, $t$ is defined between $(k-1)\tau\leq t\leq k\tau$. According to~\textbf{Assumption 1} and~\cite{Wang2019Adaptive}, we have
\begin{equation}
\theta_{[k]}^{(t)}>\varepsilon,
\end{equation}
and
\begin{equation}\label{ss}
\begin{aligned}
\frac{1}{\theta_{[k]}^{(T)}}-\frac{1}{\theta_{[1]}^{(0)}} &\geq \sum\limits_{k=1}^{K-1}\left(\frac{1}{\theta_{[k+1]}^{(k\tau)}}-\frac{1}{\theta_{[k]}^{(k\tau)}}\right)\\
&\quad+T\omega\eta\left(1-\frac{\beta\eta}{2}\right).
\end{aligned}
\end{equation}
According to the definition of $\boldsymbol{v}_{[k+1]}^{(k\tau)}$, we have
\begin{equation}
\boldsymbol{v}_{[k+1]}^{(k\tau)}=\bar{\boldsymbol{w}}^{(k\tau)}.
\end{equation}
Then, we have
\begin{equation}\label{dd}
\begin{aligned}
\frac{1}{\theta_{[k+1]}^{(k\tau)}}&-\frac{1}{\theta_{[k]}^{(k\tau)}}
=\frac{\theta_{[k]}^{(k\tau)}-\theta_{[k+1]}^{(k\tau)}}{\theta_{[k]}^{(k\tau)}\theta_{[k+1]}^{(k\tau)}}
=\frac{F\left(\boldsymbol{v}_{[k]}^{(k\tau)}\right)-F\left(\boldsymbol{v}_{[k+1]}^{(k\tau)}\right)}{\theta_{[k]}^{(k\tau)}\theta_{[k+1]}^{(k\tau)}}\\
&=\frac{F\left(\boldsymbol{v}_{[k]}^{(k\tau)}\right)-F\left(\bar{\boldsymbol{w}}^{(k\tau)}\right)}{\theta_{[k]}^{(k\tau)}\theta_{[k+1]}^{(k\tau)}}
\geq -\frac{\rho\Delta}{\varepsilon^{2}}.
\end{aligned}
\end{equation}
Combining Eq.~\eqref{ss} with Eq.~\eqref{dd}, we have
\begin{equation}\label{yy}
\begin{aligned}
\frac{1}{\theta_{[k]}^{(T)}}-\frac{1}{\theta_{[1]}^{(0)}}\geq &-\frac{\rho(K-1)\Delta}{\varepsilon^{2}}
+T\omega\eta\left(1-\frac{\beta\eta}{2}\right).
\end{aligned}
\end{equation}
When $k=K$, according to the~\textbf{Assumption 1}, we can obtain
\begin{equation}\label{ep}
F(\widehat{\boldsymbol{w}}^{(T)})-F(\boldsymbol{w}^{*})\geq\varepsilon.
\end{equation}
Therefore, Eq.~\eqref{ep} can be expressed as
\begin{equation}\label{tt}
\begin{aligned}
&\frac{1}{F\left(\bar{\boldsymbol{w}}^{(T)}\right)-F\left(\boldsymbol{w}^{*}\right)}-\frac{1}{\theta_{[K]}^{(T)}}\\
&=\frac{\theta_{[K]}^{(T)}+F\left(\boldsymbol{w}^{*}\right)-F\left(\bar{\boldsymbol{w}}^{(T)}\right)}{\left(F(\bar{\boldsymbol{w}}^{(T)})-F(\boldsymbol{w}^{*})\right)\theta_{[K]}^{(T)}}\\
&=\frac{F(\boldsymbol{v}_{[k]}^{(T)})-F(\bar{\boldsymbol{w}}^{(T)})}{\left(F(\bar{\boldsymbol{w}}^{(T)})-F(\boldsymbol{w}^{*})\right)\theta_{[K]}^{(T)}}\geq\frac{-\rho\Delta}{\varepsilon^{2}}.
\end{aligned}
\end{equation}
Summing up equation (\ref{yy}) and (\ref{tt}), we have
\begin{equation}
\begin{aligned}
&\frac{1}{F(\bar{\boldsymbol{w}}^{(T)})-F(\boldsymbol{w}^{*})}-\frac{1}{\theta_{[1]}^{(0)}}\\
&=T\omega\eta\left(1-\frac{\beta\eta}{2}\right)-\frac{{\rho K\Delta }}{{{\varepsilon ^2}}}\\
&\overset{(a)}=T\left(\omega\eta\left(1-\frac{\beta\eta}{2}\right)-\frac{\rho \Delta}{\tau\varepsilon^{2}}\right),
\end{aligned}
\end{equation}
where step (a) is obtained by $T=K\tau$.

Note that $\theta_{[1]}^{(0)}>0$, the above inequality can be simplified as
\begin{equation}\label{reciprocal}
\begin{aligned}
&\frac{1}{F(\widehat{\boldsymbol{w}}^{(T)})-F(\boldsymbol{w}^{*})}
\geq T\left(\omega\eta(1-\frac{\beta\eta}{2})-\frac{\rho \Delta}{\tau\varepsilon^{2}}\right).
\end{aligned}
\end{equation}
According to the definition we know that $\boldsymbol{w}^{*}$ is the optimal model parameters minimizing the $F(\boldsymbol{w})$. Hence $F(\widehat{\boldsymbol{w}}_{[k]}^{(T)})\geq F(\boldsymbol{w}^{*})$ and this inequality can be true when the right-hand side of the inequality is less than $0$.
Note that when the upper bound of parameter $\Theta$ or the unreliable $p_U$ or the additive noise power $\sigma$ is big enough, $T\left(\omega\eta(1-\frac{\beta\eta}{2})-\frac{\rho \left(\phi(\tau)+\frac{{ {p_\textrm{U}}}}{M}\left[ {(1 - \alpha )M\Theta  + \frac{{2\sqrt M \sigma }}{\pi }} \right]\right)}{\tau\varepsilon^{2}}\right)$ will less than zero. In this case, although the inequality (\ref{reciprocal}) is true or not true it will make no any sense. This can be interpreted as that the system will crash when the learning circumstance is unacceptable. Similarly, when local training epoch $\tau$ continues to increase  without limitation, the right-hand side of this inequality will be smaller than zero. Therefore, for ease of analysis we assume there are certain limits on $\Theta$, $p_U$, $\sigma$ and $\tau$. In other words we assume that $T\left(\omega\eta(1-\frac{\beta\eta}{2})-\frac{\rho \Delta}{\tau\varepsilon^{2}}\right)>0$. Then taking the reciprocal of the
above inequality yields
\begin{equation}\label{reverse bound}
\begin{aligned}
&F(\widehat{\boldsymbol{w}}^{(T)})-F(\boldsymbol{w}^{*})\\
&\leq\frac{1}{T\left(\omega\eta(1-\frac{\beta\eta}{2})-\frac{\rho \left(\phi(\tau)+\frac{{ {p_\textrm{U}}}}{M}\left[ {(1 - \alpha )M\Theta  + \frac{{2\sqrt M \sigma }}{\pi }} \right]\right)}{\tau\varepsilon^{2}}\right)}.\\
\end{aligned}
\end{equation}
This completes the proof. $\hfill\square$
\section{Proof of Proposition~\ref{proposition:optim_tau}}\label{appendix:optim_tau}
First, we define $H(\tau)$ as
\begin{equation}
H(\tau)\triangleq\frac{\phi(\tau)+\nabla}{\tau},
\end{equation}
where $\phi(\tau)=\frac{\delta}{\beta}\left((\eta\beta+1)^{\tau}-1\right)-\eta\delta \tau$ and $\nabla=\frac{p_{\textrm{U}}}{M}\left[ {(1 - \alpha )M\Theta  + \frac{{2\sqrt M \sigma }}{\pi }} \right] > 0$, respectively.

With a slight abuse of $\tau$, we consider continuous values of $\tau>1$, and then have
\begin{equation}
\begin{aligned}
H'(\tau)=-\frac{\nabla}{\tau^{2}}
+\frac{\delta\left(\eta\beta+1)^{\tau}(\ln(\eta\beta+1)^{\tau}-1\right)+\delta}{\beta\tau^{2}},
\end{aligned}
\end{equation}
and
\begin{equation}\label{ha}
\begin{aligned}
&H''(\tau)\\
&=\frac{\delta}{\beta\tau^{4}}\left(\tau^{3}(\eta\beta+1)^{\tau}\ln^{2}(\eta\beta+1)\right.\\
&\quad\left.-2\tau(\eta\beta+1)^{\tau}(\ln(\eta\beta+1)^{\tau}-1)-2\tau\right)+\frac{2\nabla}{{\tau}^3}\\
&=\frac{\delta}{\beta\tau^{3}}\left((\eta\beta+1)^{\tau}((\ln(\eta\beta+1)^{\tau}-1)^{2}+1)-2\right)+\frac{2\nabla}{{\tau}^3}.\\
\end{aligned}
\end{equation}
We then define $f(x)$ as
\begin{equation}\label{gg}
f(x)\triangleq x\left((\ln x-1)^{2}+1\right)-2,
\end{equation}
then, the derivate of $f(x)$ can be expressed as
\begin{equation}
\begin{aligned}
f'(x)&=(\ln x-1)^{2}+2(\ln x-1)+1\\
&=(\ln x-1+1)^{2}\\
&=\ln^{2}x\geq 0.
\end{aligned}
\end{equation}
Note that $(\eta\beta+1)^{\tau}\geq 1$ since $\tau\geq 1, \eta\beta\geq 0$, we can know that
\begin{equation}\label{result}
f(\eta\beta+1)^{\tau}\geq f(1)=0.
\end{equation}
Combining~\eqref{ha},~\eqref{gg} and \eqref{result}, we have
\begin{equation}
H''(\tau)\geq 0.
\end{equation}
Therefore, under the unreliable behaviors of clients with a fixed T, the convergence upper-bound is a convex function of the number of local epochs $\tau$.
This completes the proof. $\hfill\square$
%
\bibliographystyle{IEEEtran}
\bibliography{reference}
\end{document}